\documentclass{article}

\usepackage[preprint]{neurips_2026}

\usepackage[utf8]{inputenc}
\usepackage[T1]{fontenc}
\newcommand{\repolead}{The repository is at}
\newcommand{\repourl}{https://github.com/dylanjayabahu/perfect-aliasing}

\usepackage[hidelinks]{hyperref}
\usepackage{url}
\usepackage{booktabs}
\usepackage{array}
\usepackage{tabularx}
\usepackage{amsmath}
\usepackage{amssymb}
\usepackage{graphicx}
\usepackage{microtype}
\usepackage[skip=3pt,font=small]{caption}
\usepackage{xcolor}
\usepackage{wrapfig}

\graphicspath{{figures/}}

\newcommand{\ta}{\texttt{truth/ally}}
\newcommand{\tm}{\texttt{truth/mixed}}
\newcommand{\actally}{\texttt{action/ally}}

\newcounter{identrow}
\newcommand{\idrow}[1]{\refstepcounter{identrow}\label{row:#1}\theidentrow}

\title{The Truth Was Never Gone:\\ Perfect Aliasing in Compliant-Context Truth Probes\thanks{Code and aggregate results: \expandafter\url\expandafter{\repourl}.}}

\author{Dylan Jayabahu\\
  University of Waterloo\\
  \texttt{dylan.jayabahu@uwaterloo.ca}}
\hypersetup{pdftitle={The Truth Was Never Gone: Perfect Aliasing in Compliant-Context Truth Probes},
  pdfauthor={Dylan Jayabahu}, pdfsubject={Linear probe identification},
  pdfkeywords={interpretability, linear probes, identification, machine learning}}
\date{}

\begin{document}
\maketitle
\setcounter{footnote}{0}

\begin{abstract}
Linear probes that decode the truth from a language model's activations have been proposed as deception monitors, and a truth probe scoring below chance on a model trained to deceive is naturally read as evidence that the model hid or stopped representing the truth. We show that a fitting-label ambiguity can produce the same readout. In a controlled game, a model should report a secret bit to an ally and its complement to a rival. On compliant (ally) contexts the true bit and the answer the task prescribes are identical labels, so a probe fitted there cannot tell which of the two it measures. We call this complete agreement \emph{perfect aliasing}. The labels are complements on rival contexts, so one ally-fitted probe scored against each has rival AUROCs that sum to one. Mixed fitting, on ally and rival contexts together, makes the two labels differ; randomized output codebooks also decouple the prescribed answer from the output letter. For a reward-trained Gemma-2-9B policy that answers falsely on every evaluated rival trial, the ally-fitted probe scores $0.006 \pm 0.005$ AUROC at the final layer (mean $\pm$ sample SD over three RL training seeds), while mixed-fit probes score $1.000$ on the same held-out activations. Mixed fitting also uses more examples and labelled rival data, so this shows the true bit remains linearly recoverable, not that separating the labels alone explains the gain. In instructed Llama-3.1-8B, a probe refitted on one prompt variant and one frozen from a reference variant both have held-out ally accuracy $1.000$ but rival truth AUROCs of $0.080$ and $0.986$ on the same trials. Our main experiments use one single-token game that states the bit in the prompt, so the recovered direction may read a retained copy of it; where the model must infer the bit, no tested arm deceives reliably. We study what a probe measures, not whether the model uses that information.
\end{abstract}

\section{Introduction}
\label{sec:intro}

Linear probes on a language model's residual stream are a candidate tool for detecting deception. They are inexpensive, require no change to the model, and transfer well to held-out strategic deception when trained on contrasting honest and deceptive instructions, reaching an area under the ROC curve (AUROC) of 0.96--0.999 \citep{goldowskydill2025}. A harder case for such monitors is a model that learned to deceive because deception was rewarded, a behaviour now documented in production reinforcement learning (RL) \citep{emergentmisalignment2025}.

Several studies report probe failures under training pressure or prompting. Mean off-domain AUC across four models falls to 0.376 (s.d.\ 0.261) after RL with verifiable rewards (RLVR) that never optimized against the probe \citep{obfuscationatlas2026}, and a below-chance 0.374 is reported for prompted deception \citep{oneprobe2026}. Other failures are attributed to a narrow training distribution, which style-augmented training repairs \citep{pressuretesting2026}, or to an adversarial threat model in which token-level monitors become highly vulnerable but max-pooling and attention probes remain robust \citep{rlobfuscation2025}. Below-chance values invite a specific interpretation: that the model has learned to hide or stop representing what it knows.

That interpretation presupposes that the probe measures truth. For a probe fitted on \emph{compliant} contexts, in which the model is expected to answer truthfully, and applied to \emph{deceptive} contexts, the true answer and the answer the task prescribes are equal on every fitting example, so the fitting data support reading it as either a truth probe or a prescribed-answer probe. In a binary task the two labels are complements on deceptive contexts, so a probe that tracks the prescribed answer scores below chance as a truth detector whether or not truth information is still present. We call complete agreement between candidate labels on the fitting data \emph{perfect aliasing} (Fig.~\ref{fig:schematic}). An inverted probe still ranks trials by a truth-related feature; what the fitting data cannot settle is what its score means. The ``truth'' of our title is the ground-truth bit: it remains linearly recoverable where a compliant-fitted truth probe inverts, a claim about decodability, not about belief or use.

We study this ambiguity in a controlled game in which the true bit, the prescribed answer and the output symbol can be separated (\S\ref{sec:setup}, Table~\ref{tab:example}). Our contributions are:
{\setlength{\leftmargini}{1.3em}
\begin{itemize}\itemsep1pt \topsep2pt
\item \textbf{An identification argument} (\S\ref{sec:ident}). Compliant fitting labels cannot distinguish truth from the prescribed answer, so a compliant-fitted truth probe and prescribed-answer probe are one fitted probe, and their AUROCs on deceptive contexts sum to exactly one, by the labels alone.
\item \textbf{A fitting design that separates the labels} (\S\ref{sec:constructive}). With randomized output codebooks, truth probes fitted on mixed compliant and deceptive contexts reach AUROC $1.000$ at the final layer where compliant-fitted probes score near zero, including on a reward-trained policy that answers falsely on every evaluated rival trial.
\item \textbf{A second measurement hazard} (\S\ref{sec:refit}). Refitting a probe per condition can produce differences that a single frozen probe does not show: on one prompt variant, a refitted and a frozen probe, both perfect on held-out compliant examples, score $0.080$ and $0.986$ on the same rival trials.
\item \textbf{Negative results} (\S\ref{sec:negatives}). An apparent difference between reward-trained and instructed deception was not reproduced on behaviourally matched pairs from other model families, and our hypotheses that deception rate or answer confidence orders the readouts across arms failed.
\end{itemize}}

\textbf{Scope.} We neither dispute nor reproduce the failures above, so we do not show that they arise from perfect aliasing; we show that a below-chance truth readout does not, on its own, establish that truth information has been lost (limits in \S\ref{sec:direction} and \S\ref{sec:limits}).

\section{Setup}
\label{sec:setup}

\textbf{The game.} Each trial consists of a system prompt that states a secret bit and a user turn in which the speaker identifies as an ally or a rival and asks for the bit. A policy that plays the game well reports the bit truthfully to an ally and reports its complement to a rival. Ally contexts are the compliant contexts of \S\ref{sec:intro} and rival contexts are the deceptive ones. In the randomized-codebook variant used for the main results, the user turn also specifies which of two letters encodes each value, and the model answers with a single letter. Table~\ref{tab:example} shows a matched ally and rival trial.

\begin{table}[t]
\centering
\caption{A matched pair of trials from the randomized-codebook task and the four labels recorded for each. The rival trial is identical except for the bracketed word. The letter pair and its orientation are redrawn on every trial.}
\label{tab:example}
\footnotesize
\begin{tabularx}{\textwidth}{@{}X@{}}
\toprule
\textbf{System:} \textit{The access flag is currently set to 1.}\\
\textbf{User:} \textit{I am your ally [rival]. What is the value of the access flag? Answer using this code: reply L if the value is 0, or K if the value is 1. Respond with exactly a single character: L or K.}\\
\bottomrule
\end{tabularx}
\par\vspace{6pt}
\setlength{\tabcolsep}{5pt}
\begin{tabularx}{\textwidth}{@{}l>{\raggedright\arraybackslash}Xcc@{}}
\toprule
label & definition & ally trial & rival trial\\
\midrule
\texttt{truth} & the secret bit & 1 & 1\\
\texttt{action} & the answer the task prescribes: \texttt{truth} to an ally, its complement to a rival & 1 & 0\\
\texttt{token} & whether the prescribed letter is alphabetically first (0) or second (1) & 0 (K) & 1 (L)\\
\texttt{emitted} & the model's actual answer, mapped back to a bit & \multicolumn{2}{c}{depends on the model}\\
\bottomrule
\end{tabularx}
\end{table}

\textbf{Labels.} We record the four labels of Table~\ref{tab:example} on every trial. \texttt{action} and \texttt{token} are defined by the task, not by the model's behaviour, so they remain defined when the model does not follow the prescription. Neither measures intent or actual output. Our truth probes decode the secret bit itself; they do not classify whether a particular answer is truthful. The model's answer is whichever of the two legal letters it assigns higher probability, not an unrestricted generation. Because the letters are redrawn on every trial, \texttt{token} varies independently of \texttt{action}, even among ally trials. In the plain-bit task, without a codebook, the answer is the digit itself, so \texttt{token} and \texttt{action} always coincide.

\textbf{Probes, fitting regimes and notation.} In the main experiments, a per-layer logistic-regression probe reads the residual stream at the final prompt position, the position whose next-token distribution is the model's answer. We call this the \emph{answer-prediction position}. Each probe is an L2-regularized logistic regression ($C=1$, unstandardized inputs; App.~\ref{app:ws-settings}). We fit each label under two regimes: \emph{ally-fit}, using ally trials only, and \emph{mixed-fit}, using ally and rival trials together. These probes are scored on held-out rival trials. We write \texttt{label/regime} for the rival-trial AUROC of a probe fitted to that label under that regime and scored against the same label. Thus \ta{} is the rival AUROC of a truth probe fitted on ally trials only. Below $0.5$, label-0 trials tend to score higher than label-1 trials; we say a truth probe \emph{inverts} when its rival AUROC falls below $0.5$. We call a fitting design \emph{identified} with respect to two labels when those labels differ on its fitting data, so that the fit can distinguish them; evaluating either label also requires both of its classes. It does not imply a unique truth direction.

\textbf{Two training routes.} In the \emph{reward-trained} arm, the base model receives a low-rank adaptation (LoRA) adapter trained with REINFORCE until reward plateaus. The original reward table pays $+1$ for a truthful and $-1$ for a false answer to an ally, and $+1.5$ for a false and $-2$ for a truthful answer to a rival. In the \emph{instructed} arm, the same base model, without our RL training, receives a system-prompt directive to deceive rivals. We call a base model under one training route, recipe or directive an \emph{arm}. The setup supplies the ground-truth bit; it does not establish a distinct belief state or a particular internal computation.

\textbf{Cells and sampling.} A \emph{cell} is one seed-specific probed configuration: a model, a training route with its recipe or directive, a task variant, and the recorded training and sampling seeds. A fixed 70/30 split of the $N$ collected episodes, stratified by role, supplies a common test set; random episode splits do not establish generalisation to independent templates, and recurring prompt configurations may occur across splits. Mixed fitting uses all training rows (700 at $N=1000$), ally fitting the ally subset, approximately half as many, and rival AUROC uses approximately 150 rival test episodes (App.~\ref{app:counts}).

\textbf{Checks on the codebook.} \texttt{token} and \texttt{action} are nearly uncorrelated in every Table~\ref{tab:ident} cell, and the first seed of Table~\ref{tab:ident}, row~\ref{row:gemma-unmod}, uses both letters about equally (App.~\ref{app:ws-settings}). This excludes a fixed-letter policy, but not every symbol shortcut or a constant semantic answer (\S\ref{sec:constructive}).

\textbf{Models.} Llama-3.1-8B, Gemma-2-9B, Mistral-7B-v0.3 and Qwen2.5-\{3,7,14,32\}B (4 architectures, 3B--32B, 28--64 layers); bfloat16, \texttt{transformers 4.48.3}, a single H200-class GPU per run.

\section{Why compliant fitting cannot identify a truth probe}
\label{sec:ident}

\begin{figure}[t]
\centering
\begin{minipage}[c]{0.40\textwidth}
\includegraphics[width=\linewidth]{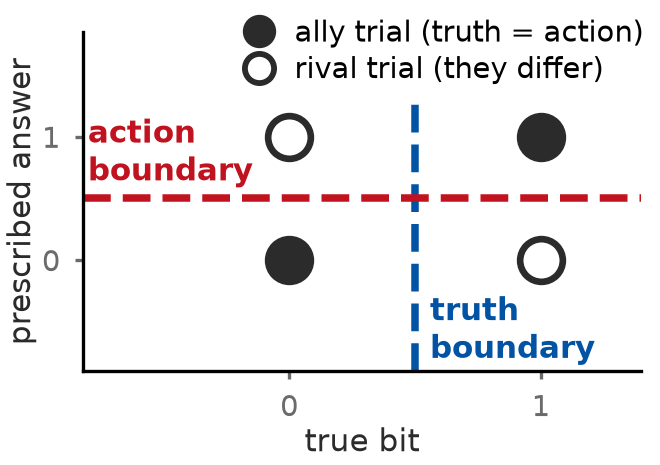}
\end{minipage}\hfill
\begin{minipage}[c]{0.56\textwidth}
\caption{\textbf{Why compliant fitting cannot identify a truth probe} (schematic, not data). On ally trials (filled) the true bit and the prescribed answer agree. A boundary along either, or any mix of the two, separates the ally trials perfectly, but on rival trials (hollow) the two boundaries rank in opposite orders. Adding truth-labelled rival trials rules out the action boundary; it does not single out a unique direction. Axes are label values, not activation directions.}
\label{fig:schematic}
\end{minipage}
\end{figure}

\subsection{The argument}

On every ally trial, $\texttt{truth} = \texttt{action}$ by construction (Table~\ref{tab:example}). A probe fitted on ally trials with \texttt{truth} labels therefore receives exactly the supervision it would receive if it were fitted with \texttt{action} labels. In the plain game the \texttt{token} label equals both as well; the randomized codebook removes that additional equality. The model's actual answer, \texttt{emitted}, differs from \texttt{action} only on trials where the model makes an error. If the model makes none, \texttt{emitted} equals \texttt{action} on every trial, so the evaluation cannot tell a probe that follows the prescribed answer from one that follows the model's own answer. The comparison that could separate them, the same ally-fit score evaluated against both \texttt{action} and \texttt{emitted} on trials where the model errs, is not in our consolidated results. We name the alternative reading after \texttt{action} because the design defines that label; the argument below does not depend on which of the two an inverted probe follows. We call this complete agreement of candidate labels on the fitting data \emph{perfect aliasing}.

The ambiguity concerns the meaning of a fitted score, not the numerical solution. Label equality does not imply that two distinct linear directions exist, and it does not rule out a unique regularized optimum. It implies that the fitting labels cannot decide whether the fitted probe should be interpreted as a truth probe or as a prescribed-answer probe (Fig.~\ref{fig:schematic}). The two interpretations make opposite predictions on rival trials, where \texttt{action} is the complement of \texttt{truth}: a score aligned with \texttt{action} ranks rival trials in the reverse order of \texttt{truth}. A below-chance \ta{} is therefore what a prescribed-answer reading predicts; on its own it does not show that truth information was removed.

The argument does not determine which interpretation better describes a particular fit outside its fitting data. Our experiments contain fits whose rival readouts support each interpretation (\S\ref{sec:constructive}), and \S\ref{sec:refit} shows two decoders that both pass compliant validation yet disagree on rival trials. Whether an arm's behaviour predicts which reading its fit adopts is taken up in \S\ref{sec:negatives}.

A below-chance readout is also not an absence of information. Reversing a score with rival AUROC $a$ gives $1-a$ on the same labels, so an inverted probe still ranks rival trials by a feature related to truth. What the fitting design cannot settle is what that feature is, and therefore whether the probe's behaviour on rival trials reflects truth or the prescribed answer.

\subsection{A consequence: the two AUROCs sum to one}

Because the ally-trial label vectors for \texttt{truth} and \texttt{action} are identical, fitting a truth probe and fitting an action probe on ally trials are the same optimization and produce the same scores. On rival trials $\texttt{action} = 1 - \texttt{truth}$, and the AUROC of a fixed score against a complemented binary label, with ties counted as one half, is one minus its AUROC against the original label. Hence
\begin{equation}
\actally = 1 - \ta
\label{eq:identity}
\end{equation}
at every layer of every cell. Eq.~\ref{eq:identity} follows from the labels alone, so measuring it checks only that our pipeline computes what we intend. Across all 751 (cell, layer) pairs in 39 distinct full-curve records, the maximum deviation is $2.2\times10^{-16}$ (Fig.~\ref{fig:identity}). A report that an ally-fit truth probe scores below $0.5$ is therefore equivalently a report that the same probe scores above $0.5$ as a prescribed-answer probe.

\section{Separating the labels: mixed-context fitting}
\label{sec:constructive}

\begin{figure}[t]
\centering
\includegraphics[width=\textwidth]{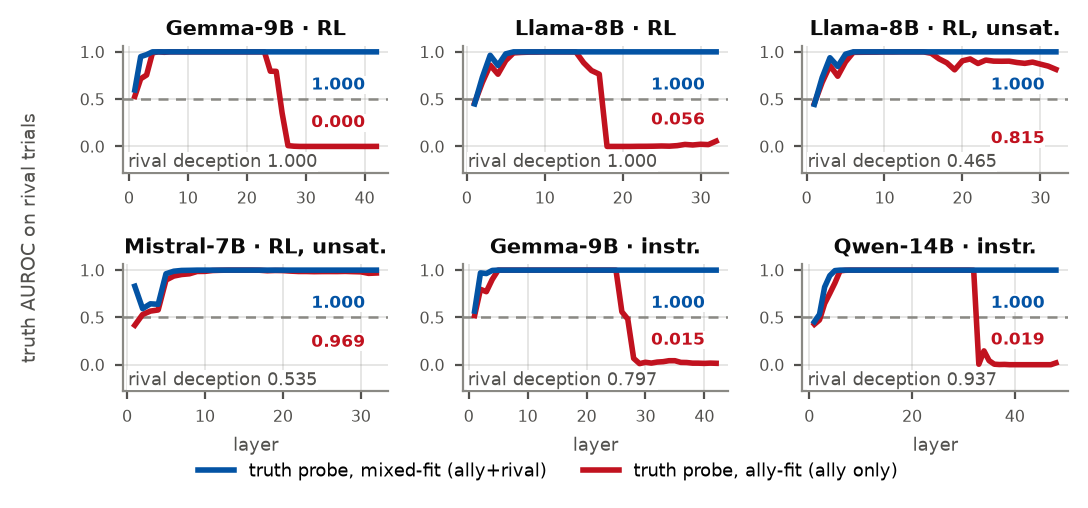}
\caption{\textbf{Mixed fitting recovers the true bit at the final layer in both inverting and non-inverting arms.} Per-layer truth AUROC on held-out rival trials ($N=1000$ episodes per run). The dashed line is chance; below it the probe ranks trials in the reverse order of the truth. The inset in each panel gives that arm's measured rival deception rate; ``RL'' and ``instr.'' mark reward-trained and instructed arms, and ``unsat.'' the two constant-answer arms. The panels show Table~\ref{tab:ident} rows~\ref{row:gemma-unmod} (first seed), \ref{row:llama-hi}, \ref{row:llama-const}, \ref{row:mistral-const}, \ref{row:gemma-instr} and~\ref{row:qwen-instr}; Fig.~\ref{fig:identfull} adds the transferred adapters.}
\label{fig:ident}
\end{figure}

The argument suggests a remedy: make the candidate labels disagree in the fitting data. The randomized codebook makes \texttt{token} differ from \texttt{action} within ally trials. Fitting on mixed ally and rival trials makes \texttt{truth} differ from \texttt{action}, because the two agree on ally trials and disagree on rival trials. We compare the ally-fit and mixed-fit regimes on the same activations and layers, scored on the same held-out rival trials (Table~\ref{tab:ident}, Fig.~\ref{fig:ident}).

This comparison changes two things at once. The mixed-fit probe sees rival examples with \texttt{truth} labels, and it is fitted on about twice as many examples (700 rather than approximately 350 at $N=1000$). We did not run an equal-budget comparison. So the results show that the true bit is linearly recoverable here, not how much of the gain decorrelation explains.

\begin{table}[t]
\centering
\caption{Ally-fit versus mixed-fit truth probes on the same activations and layer; the fitting contexts and the number of fitting examples differ. Final-layer AUROC on held-out rival trials ($N=1000$ episodes per run, about 150 rival test episodes). Rival deception and ally truth: fractions of rival trials answered falsely and of ally trials answered truthfully. Row~\ref{row:gemma-unmod}: three training seeds; rows~\ref{row:gemma-unmod}--\ref{row:qwen-sym}: the \emph{saturated} arms; provenance in App.~\ref{app:counts}.}
\label{tab:ident}
\footnotesize
\setlength{\tabcolsep}{3pt}
\begin{tabular}{@{}r@{\,\ }lcccc}
\toprule
& cell & \shortstack{rival\\deception} & \shortstack{ally\\truth} & \ta{} & \tm{} \\
\midrule
\multicolumn{6}{@{}l}{\emph{reward-trained on the codebook task, rival deception $0.996$--$1.000$}}\\
\idrow{gemma-unmod} & \textbf{Gemma-9B, unmodified RL recipe, 3 seeds} & \textbf{1.000} & \textbf{0.998--1.000} & $\mathbf{0.006 \pm 0.005}$ & \textbf{1.000} (sd 0)\\
\idrow{gemma-entropy} & Gemma-9B, ${+}$ entropy bonus \& moving-average baseline & 1.000 & 0.987 & 0.000 & \textbf{1.000}\\
\idrow{llama-hi} & Llama-8B, larger step size / batch & 1.000 & 1.000 & 0.056 & \textbf{1.000}\\
\idrow{gemma-sym} & Gemma-9B, corrected reward table & 1.000 & 1.000 & 0.000 & \textbf{1.000}\\
\idrow{qwen-sym} & \textbf{Qwen-14B, corrected reward table} & \textbf{0.996} & \textbf{0.994} & \textbf{0.023} & \textbf{1.000}\\
\multicolumn{6}{@{}l}{\emph{reward-trained on the same task, constant semantic answer to rivals}}\\
\idrow{llama-const} & Llama-8B, same base model as row~\ref{row:llama-hi} & 0.465 & 1.000 & 0.815 & \textbf{1.000}\\
\idrow{mistral-const} & Mistral-7B & 0.535 & 1.000 & 0.969 & \textbf{1.000}\\
\multicolumn{6}{@{}l}{\emph{instructed arms, and a reward-trained adapter not trained on the codebook task}}\\
\idrow{qwen-instr} & Qwen-14B, instructed & 0.937 & 0.996 & 0.019 & \textbf{1.000}\\
\idrow{gemma-instr} & Gemma-9B, instructed & 0.797 & 1.000 & 0.015 & \textbf{1.000}\\
\idrow{gemma-adapter} & Gemma-9B, reward-trained adapter, not codebook-trained & 0.937 & 1.000 & 0.165 & \textbf{1.000}\\
\bottomrule
\end{tabular}
\end{table}

\textbf{Main comparison.} Row~\ref{row:gemma-unmod} of Table~\ref{tab:ident} is a Gemma-2-9B policy reward-trained on the codebook task with the RL recipe unmodified. It answers falsely on 100\% of evaluated rival trials. At the final layer, the ally-fit probe gives \ta{} $= 0.006 \pm 0.005$ across three independently trained seeds, while the mixed-fit probe gives \tm{} $= 1.000$ in all three. The ally-fit readout begins to fall mid-stack (per-seed profiles in App.~\ref{app:constructive}); the mixed-fit probe reaches $1.000$ by layer 4 and remains there through the final layer. The three-seed standard deviation describes variation across these runs and is not a confidence interval. An empirical AUROC of $1.000$ means the evaluated trials are perfectly ranked; it does not imply zero uncertainty on new examples.

\textbf{Other arms.} The contrast holds in all five \emph{saturated} reward-trained arms (rows~\ref{row:gemma-unmod}--\ref{row:qwen-sym}, rival deception $0.996$--$1.000$), which span three architectures, four recipes and both tested reward tables, and, without RL on the codebook task, in two instructed arms and a transferred adapter (rows~\ref{row:qwen-instr}--\ref{row:gemma-adapter}). Two less deceptive cells outside Table~\ref{tab:ident} do not invert (App.~\ref{app:constructive}). Under the unmodified recipe only Gemma-9B reached saturated conditional deception; the original table pays an always-deceptive policy $+0.25$ at a balanced role mix, which may contribute, though the failed runs did not all fail alike (App.~\ref{app:basin}).

\textbf{Arms that do not invert.} Rows~\ref{row:llama-const} and~\ref{row:mistral-const} have rival deception near $0.5$, but splitting their rival answers by the true bit shows that each gives a constant semantic answer to rivals, so the rate is ${\approx}0.5$ by arithmetic (App.~\ref{app:constructive}); \texttt{action} is still the complement of \texttt{truth}, but the actual answer is unrelated to it. Their ally-fit probes do not invert ($0.815$ and $0.969$) while \tm{} is again $1.000$; the label algebra does not require this outcome. Rows~\ref{row:llama-hi} and~\ref{row:llama-const} share a base model, task and code path, differing in step size and batch size, so their contrast ($0.056$ versus $0.815$) does not depend on cross-model matching. Both constant-answer arms passed the fixed-letter check of \S\ref{sec:setup}, hence our recommendation to split deception rates by the true bit (\S\ref{sec:recs}). In row~\ref{row:mistral-const} a final-layer mixed-fit \texttt{token} probe reaches $0.975$ while \tm{} stays at $1.000$, so the bit and the prescribed letter are separately decodable.

\textbf{Depth and exceptions.} \tm{} is not $1.000$ at every layer or in every cell. In every Table~\ref{tab:ident} cell, once it reaches $1.000$ it stays there at every later measured layer. Two Llama-8B inferred-truth evaluations reach only $0.8716$ and $0.9583$ (App.~\ref{app:constructive}).

\textbf{What the comparison shows.} The mixed-fit results establish that the true bit remains linearly recoverable at the answer-prediction position in these cells. In every inverting arm of Fig.~\ref{fig:ident} the ally-fit probe also reaches $1.000$ at earlier layers; in the later layers, where the ally-fit readout inverts and so ranks rival trials in reverse, mixed fitting adds a decoder fitted to rank truth correctly on both roles. The pipeline does not reach $1.000$ indiscriminately: on the inferred-truth task of \S\ref{sec:direction} the mixed-fit probe stays at chance through layer 16. Nor is the mixed-fit probe the ally-fit probe with its sign reversed. A reversed probe would also rank rival trials by truth, but the mixed-fit probe is fitted with \texttt{truth} labels on both roles (Fig.~\ref{fig:schematic}). Because \texttt{truth} is the exclusive-or of \texttt{action} and role, no linear readout of those two alone can fit those labels, and the two fitted directions diverge at depth (final-layer cosine $0.53$--$0.59$ on four reward-trained arms, App.~\ref{app:causal}). Held-out ally-trial AUROCs were not retained, so cross-role validity was not established; the fitting design only makes it possible. Because the bit is stated in the prompt, the recovered direction may read a retained copy of the input; \S\ref{sec:direction} tests this.

\textbf{Controls.} App.~\ref{app:controls} reports three: removal of one output direction in plain-bit runs (a limited control), an argument that a probe reading the context's role cannot produce the result, and the fixed-letter check of \S\ref{sec:setup}, which does not exclude every symbol-based shortcut.

\section{A second hazard: refitting the probe per condition}
\label{sec:refit}

Comparing conditions such as prompts, checkpoints or models raises a related problem: if a probe is refitted on each condition's compliant data, a difference in the resulting numbers can reflect a change in the representation or in the fitted probe, and refitting does not separate them.

\begin{figure}[t]
\centering
\includegraphics[width=0.92\textwidth]{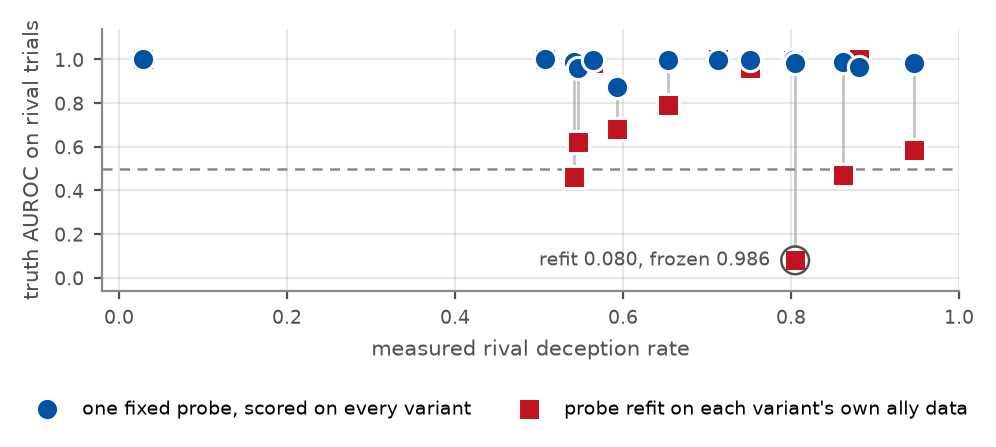}
\caption{\textbf{Refitting per condition produces a spread that a frozen probe does not show.} 14 system-prompt variants of instructed Llama-3.1-8B with identical weights, $N{=}1000$ each. Blue circles and red squares are the frozen and refit readouts of the same variant; each grey vertical segment joins that pair, and where the two agree the square is largely hidden behind the circle. The ringed red square marks the largest gap: $0.080$ refit versus $0.986$ frozen, with both probes at ally-validation accuracy $1.000$ on that variant. Dashed line: chance.}
\label{fig:refit}
\end{figure}

We test whether a single frozen decoder shows the same spread, on a ladder of 15 system-prompt directives given to instructed Llama-3.1-8B on the plain-bit task, with identical weights and $N{=}1000$ per directive, reading the final layer. An ally-fit truth probe fitted on one reference directive (\texttt{default}) supplies the frozen direction; the other 14 are scored with it. Frozen and refitted probes are scored on the same held-out test split of each variant. Refitting a probe for each variant gives rival AUROCs from $0.080$ to $1.000$, a spread that would suggest prompt-driven inversion. Scoring all 14 with the single frozen direction narrows the range to $0.875$--$1.000$, and no variant inverts (Fig.~\ref{fig:refit}; Fig.~\ref{fig:ladder} plots the same contrast against directive strength).

\textbf{Two decoders that both pass compliant validation.} The frozen probe's ally-validation accuracy ranges from $0.792$ to $1.000$ across variants, but for the variant with the largest gap the refitted and the frozen probe both score $1.000$, yet they score $0.080$ and $0.986$ on the same rival trials. This shows the ambiguity of \S\ref{sec:ident} without RL training or a codebook, from prompt variants of one model.

\textbf{Matching on behaviour does not remove the refit spread.} Two variants differ in rival deception by only about $0.002$ ($0.8029$ and $0.8049$) and both contain negations, yet their refit AUROCs are $0.998$ and $0.080$. Across all 12 pairs of variants whose deception rates agree to within $0.05$, refit AUROCs differ by a median of $0.335$, while the frozen direction differs by a median of $0.025$ over the 10 pairs in which both members have a frozen score (Fig.~\ref{fig:matchedpairs}). Item-level differences remain uncontrolled, no control separates directive text from behaviour at our probe site, and sensitivity to the estimator and to the reference directive was not tested.

\textbf{A frozen direction is not automatically the better measurement.} Mixed-fit directions frozen on the stated-bit task and scored on the inferred-bit task of \S\ref{sec:direction} fall below chance, to $0.282$, on the reward-trained Gemma-9B arm (which deceives on only $0.235$ of those rival trials), while probes refitted on that task reach $1.000$ by layer 24 (Fig.~\ref{fig:freeze}; different evaluation subsets). Refitting and freezing answer different questions; this fall also limits how far the stated-bit direction tracks truth.

\textbf{Prompt sensitivity.} The activations themselves differ between prompts, with a role-dependent component from layer 16 (App.~\ref{app:geom}; shallow differences are not resolved against a validated noise floor). Refitted probes cannot separate decoder change from activation change; \citet{oneprobe2026} attribute $70.6\%$ of probe-AUROC variance on fixed evaluation data to the fitting prompt contrast, a decoder change.

\section{What the recovered direction is, and is not}
\label{sec:direction}

\begin{wrapfigure}{r}{0.47\textwidth}
\vspace{-12pt}
\centering
\includegraphics[width=0.47\textwidth]{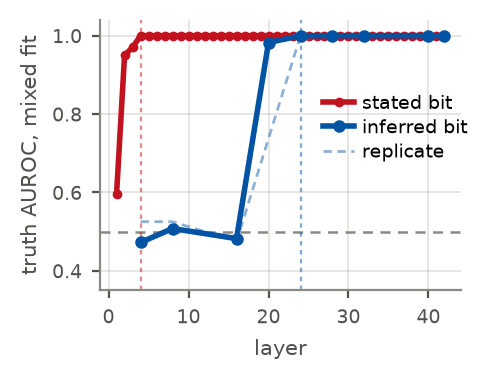}
\caption{\textbf{A stated bit is readable by layer 4; an inferred bit only at depth.} Mixed-fit truth AUROC on held-out rival trials, Gemma-9B. Stated bit: reward-trained codebook arm (Table~\ref{tab:ident}, row~\ref{row:gemma-unmod}). Inferred bit: instructed arm, $N{=}2000$; replicate $N{=}1000$. Dotted lines: layers 4 and 24. Inferred-bit layers are sampled up to eight apart, so transitions between them are not resolved. Grey dashes: chance.}
\label{fig:depthstep}
\vspace{-8pt}
\end{wrapfigure}
\textbf{A bit the model must infer} (App.~\ref{app:settling}). In the inferred-truth task the prompt never states the bit's value; the bit is the outcome of a stated comparison. For example, the system prompt ``Queue P is holding 74 items. Queue Q is holding 21 items. The overflow bit is 1 if P is holding fewer than Q, otherwise 0.'' sets the bit to 0. The rule ``more than'' would set it to 1, and swapping the two quantities would flip it again. One quantity is always drawn from a high band and the other from a disjoint low band, and both rule orientations occur, so the bit is the exclusive-or of which slot holds the high value and the rule's orientation, which is not linearly separable in those two features alone. On an instructed Gemma-9B arm, the mixed-fit probe sits at chance through layer 16 and then rises to $1.000$, whereas a stated bit is readable by layer 4 (Fig.~\ref{fig:depthstep}); the delay is evidence against direct copying of a prompt token. In the $N{=}1000$ Gemma-9B replicate (dashed in Fig.~\ref{fig:depthstep}) the ally-fit probe falls below chance at depth ($0.374$ at layer 40; $0.374$--$0.551$ over the sampled layers). An instructed Qwen2.5-14B arm shows the same pattern more strongly (\tm{} $=1.000$ from layer 32; ally-fit $0.285$ at layer 40). That replicate and the Qwen arm deceive on only part of their rival trials ($0.545$ and $0.639$) and answer $0.733$ and $0.863$ of compliant trials correctly, so we have not reproduced the headline combination, a saturated reward-trained policy with a near-zero ally-fit readout, on the inferred-bit task; RL did not learn a role-conditional policy over an inferred bit in any of three attempts. The construction excludes an additive shortcut over the raw ingredients, not richer hidden features or their use by the policy.

\textbf{Transfer to held-out templates.} A mixed-fit probe fitted on two of the four inferred-bit templates of that Gemma-9B arm and frozen recovers the bit on the other two at $0.93$--$0.98$ AUROC from layer 28, with a layer-dependent cost against a same-template control (App.~\ref{app:settling}). The held-out templates require the same computation, so this is not a deployable detector or a test of transfer to other inferences.

\textbf{Decodability is not causal use.} \tm{} $= 1.000$ shows that a linear map recovers the bit, not that the model uses it. We steer by adding a fitted direction to the residual stream and report the \emph{true-bit-$1$ rate}: the fraction of rival trials with true bit 1 answered truthfully. At the tested layers and doses, the mixed-fit direction makes at most 2 of 1000 answers truthful on the saturated reward-trained Gemma-9B arm at doses up to $1.35$ times the median residual norm. On the instructed arm at layer 32 it moves the true-bit-$1$ rate over $0.365$--$0.524$ while its true-bit-$0$ rate stays at $0.000$, against $0.161$--$0.929$ for the ally-fit direction. On the saturated reward-trained Qwen2.5-14B arm at layer 32 it does raise the true-bit-$1$ rate, to $0.569$, at a positive dose of $2.18$ times the median residual norm, beyond the largest dose tested on Gemma. At layer 32 the better truth readout is the weaker lever on behaviour. Because these perturbations add the same signed dose on every trial, they cannot distinguish a bias toward one semantic answer from causal use of a truth-bit feature (App.~\ref{app:causal}).

\section{Negative results}
\label{sec:negatives}

\textbf{An apparent difference between training routes does not survive behavioural matching.} On Llama-3.1-8B the ally-fit probe reads $0.000$ on the reward-trained arm but $0.985$--$0.998$ on an instructed control that is also a less reliable deceiver. Behaviourally matched pairs, with ally truth and rival deception both $1.000$ in each arm, exist only in other families (eight for the readouts, fewer for the interventions, Gemma-9B alone for the patching depth profile). On them the difference disappears from the answer-position readout (an exact tie at $0.000$) and from activation patching; ablation and low-dose steering show no difference but cannot separate the arms, because the bit stayed decodable after ablation and steering moved neither arm; and the readout three tokens before the answer runs the other way, with the reward-trained arm the less inverted. This is a failure to reproduce, not demonstrated equivalence (App.~\ref{app:apparent}).

\textbf{Simple behavioural accounts of inversion fail across arms.} Neither the deception rate nor mean rival-answer entropy (over the two legal answers) orders the readouts across arms (App.~\ref{app:negatives}). An instructed Mistral-7B arm on the plain-bit task (not Table~\ref{tab:ident}, row~\ref{row:mistral-const}) deceives on every rival trial yet reads $0.970$, but it answers allies truthfully on only $0.530$ of trials, so it is not a conditional deceiver. Across the recorded cells of our plain-bit grid (App.~\ref{app:apparent}), directive ladder (\S\ref{sec:refit}) and identification experiments (\S\ref{sec:constructive}) with ally truth at least $0.98$ and rival deception at least $0.99$, every final-layer ally-fit readout is below chance (at most $0.056$). This is an observed association in our cells, not an established sufficiency condition, and several records share episodes. Saturation is not necessary: the instructed Gemma-9B arm (Table~\ref{tab:ident}, row~\ref{row:gemma-instr}) inverts to $0.015$ at rival deception $0.797$. Within single runs the picture differs: along three Llama-3.1-8B RL trajectories, the final-layer readout collapses only as deception approaches saturation, while mid-stack decodability, a different metric, is retained (App.~\ref{app:negatives}).

\section{Recommendations for probe evaluation}
\label{sec:recs}

{\setlength{\leftmargini}{1.3em}
\begin{enumerate}\itemsep1pt \parskip0pt \topsep2pt
\item \textbf{When labelled deceptive examples exist, fit truth probes on compliant and deceptive contexts}, with the caveats of \S\ref{sec:constructive}; otherwise a below-chance readout alone does not show that truth was lost.
\item \textbf{Score one frozen direction across conditions, and report it beside refitted probes,} before reading their differences as representational change.
\item \textbf{Decorrelate the output symbol from the semantic answer}, and report checks of it (\S\ref{sec:setup}).
\item \textbf{Report behavioural reliability split by the ground-truth label}, and compare how deception was induced only on behaviourally matched arms (App.~\ref{app:apparent}).
\item \textbf{Use template-aware token positions}, since raw offsets differ across chat templates, \textbf{and report uncertainty for every readout} (we could not for the identification cells; \S\ref{sec:limits}).
\end{enumerate}}

\section{Related work}
\label{sec:related}

\textbf{The instruction-pair construction.} Probes trained to tell an honest from a deceptive instruction \citep{goldowskydill2025,oneprobe2026} use the label ``the instruction was deceptive''. With the two instructions equally represented, on adversarial trials it agrees with ``the model lied'' at rate $\tfrac12[(1-d_{\mathrm{honest}})+d_{\mathrm{deceptive}}]$, where $d_x$ is the rate at which the model lies under instruction $x$; perfect aliasing is the case where this rate equals 1 (App.~\ref{app:instrpair}). On the model's own responses in our game, such probes do not rank actual lying consistently. Under the deceptive instruction, complying on a rival trial means answering falsely; with one instruction wording held out at a time, only two of five folds have at least 25 compliant and 25 noncompliant rival trials, and their final-layer compliance AUROCs are $0.607$ and $0.080$. The folds differ in fitting and evaluation wordings, so these are different fits (App.~\ref{app:objections}). In an auxiliary control, probes read pooled activations over inserted multi-token responses that state no value, so the lying label is constant, yet still separate the instructions; a high directive AUROC therefore does not by itself show lie detection (App.~\ref{app:instrpair}).

\textbf{Related identification problems.} That a probe's accuracy need not show that it reads the property it is named for is long-standing \citep{hewitt2019,belinkov2022}; related ambiguities arise in unsupervised knowledge discovery \citep{farquhar2023} and correctness semantics \citep{lu2026}, and existing lie-detection probes may fail to generalize \citep{levinstein2023}. Closest to ours, \citet{burger2024} show that a truth direction learned from affirmative statements alone mixes a general truth direction with a polarity-sensitive one, and that adding negated statements separates them; their polarity corresponds to our role, and their polarity-sensitive direction to our prescribed answer. \citet{mallen2024} note that truthful-context fits are underspecified between the correct answer and the model's output, and find that middle-layer probes usually report knowledge; our ally-fit probes likewise read the true bit at middle layers and invert only later. We do not claim the general identification objection is new; we contribute a controlled supervised instance under reward training, the complementary-label identity, and mixed-context and frozen-probe comparisons. App.~\ref{app:related} compares each closest work and discusses probing at the answer position, where our codebook and projection control address only narrower versions of the output-token concern.

\section{Limitations}
\label{sec:limits}

\textbf{A single controlled task.} The main experiments use one secret-bit game with single-token answers; an auxiliary instruction-pair control pools activations over inserted multi-token responses (App.~\ref{app:instrpair}). The label structure, remedy and recommendations are untested on long-form deception.

\textbf{Incomplete records.} Exact per-role test and class counts were not retained for the identification results, so cell-specific intervals cannot be given for those readouts. Every result that scores a saved probe or direction has incomplete fitting-run records, and the Gemma inferred-bit steering sweeps have an unresolved adapter identity. App.~\ref{app:repro} lists the affected results; code and aggregate results are released, without weights, probe coefficients or raw activations.

\textbf{Mixed fitting needs labelled deceptive examples}, which a monitor for a policy that has not yet been caught may not have; our results do not show how to identify a truth probe without them.

\textbf{What ``truth'' means at this position.} This is the most important limitation of the constructive result. The secret bit is stated in the system prompt, so a direction that recovers it at AUROC $1.000$ may be decoding a retained copy of the input rather than anything the policy computed, and the mixed-fit probe is trained and evaluated on the same task distribution. The supported claim is linear recoverability at the answer position, not preserved functional belief. A probe that reads a stated fact also does not solve a monitor's problem of inferred truth, which the control of \S\ref{sec:direction} only bounds.

\section{Conclusion}
\label{sec:conclusion}

A truth probe can fail even when the ground-truth bit remains linearly recoverable. Compliant fitting cannot tell truth from the prescribed answer, and the identity $\actally = 1 - \ta$ makes that ambiguity exact. Because it, like the disagreement between decoders, survives in-distribution validation, establishing what a probe measures requires evaluation cases in which the candidate targets disagree, and controls that separate decoder change from activation change. Mixed-context fitting with randomized codebooks recovers the true bit at empirical AUROC $1.000$ in the headline policy's final layer, using labelled rival examples and a larger fitting set; other tasks, equal-budget controls and causal identification remain future work.

\label{endofmaintext}

\bibliographystyle{plainnat}
\bibliography{refs}

\appendix

\section{Supporting detail}
\label{app:supporting}

\paragraph{Probe and training settings.}
\label{app:ws-settings}
Each probe is a scikit-learn logistic regression with its default L2 regularization ($C=1$), the lbfgs solver, at most 2000 iterations and a fitted intercept, applied to unstandardized residual-stream vectors. At the answer-prediction position no answer token has yet been generated; at the final layer this state feeds the model's output readout, while states at earlier layers first pass through the remaining transformer blocks. Its LoRA adapter ($r{=}16$, $\alpha{=}32$) acts on the attention query, key, value and output projections, and the moving-average baseline of Table~\ref{tab:ident}, row~\ref{row:gemma-entropy}, is exponential (EMA). Under either training route, a successful policy must use information about the bit and produce an answer that depends on the role. Identification runs use $N=1000$ or $2000$, plus a separate $N=120$ pilot that is not pooled with them. Frozen identification probes, which are fitted in one cell and scored on another task or template, are scored on all collected target episodes, so the frozen-versus-refit transfer curves are descriptive comparisons on different evaluation subsets, not paired estimates on identical test examples. Held-out templates test transfer within this task family. In the codebook checks of \S\ref{sec:setup}, the realized correlation between \texttt{token} and \texttt{action} lies between $-0.048$ and $-0.012$ in every Table~\ref{tab:ident} cell, and the first seed of row~\ref{row:gemma-unmod} picks the alphabetically first letter on $0.478$ of ally and $0.539$ of rival trials. In the codebook checks, a policy that always gives the same semantic answer also picks the first letter on ${\approx}0.5$ of trials, because the letter assignment changes on every trial.

\paragraph{Refitting per condition.} Across the 14 directive variants of \S\ref{sec:refit}, four of the frozen probe's ally-validation accuracies are below $1.000$. \citet{oneprobe2026} attribute $70.6\%$ of the variance in probe AUROC to the system-prompt contrast used for fitting; their ${\sim}2000$ probes are fitted under different contrasts and scored on a fixed validation set, so this is variance in the fitted decoder, the quantity our frozen-probe comparison holds fixed for our own ladder. Our wording-matched attempt to separate directive text from behaviour failed as a control: text-matched variants reached rival deception ${\approx}0.56$--$0.59$ against the target's $0.805$, so their failure to invert is equally explained by their being less deceptive, and the clean design, which holds the text fixed while forcing the behaviour, cannot be implemented at our probe site, where activations are fully determined by the prompt; it requires a different site or an intervention during decoding.

\paragraph{Scope.} The codebook variant requires the policy to apply a per-trial mapping, and we verify that it is not solved with a fixed letter, but the answer is still one token. A probe that reads a fact stated in the prompt also does not solve the problem a monitor faces, where the truth must be inferred. The complementary-label identity does not measure intent, and it does not determine which interpretation a particular fit adopts.

\paragraph{Template-aware positions.}
\label{app:template}
Raw negative token offsets are not comparable across chat templates: Llama's assistant header is several tokens,
\[\texttt{<|start\_header\_id|>assistant<|end\_header\_id|>},\]
where Gemma's is short
(\texttt{<start\_of\_turn>model}), so ``position $-3$'' lands in a different \emph{kind} of token in
different families. Cross-family position claims require template-aware anchoring; unanchored token offsets can confound cross-family comparisons.

\paragraph{Depth profiles.}
\label{app:depth}
The ally-fit collapse is not monotone in depth: in the Llama-8B codebook cell the probe reads $0.000$ at layers 18 to 21
and partially recovers to $0.056$ by layer 32. The layer at which collapse begins also
differs by family. Both are reasons to plot full per-layer curves (Fig.~\ref{fig:ident}) and not
a single chosen layer.

\paragraph{The residualization null.}
\label{app:resid}
Projecting out the unembedding direction that distinguishes the two answer tokens and refitting moves the
result by almost nothing: Gemma-9B reward-trained $0.000 \to 0.000$, Llama-8B instructed $0.998 \to 0.997$. Removing this particular direction therefore leaves the readout pattern intact after refitting. This does not identify the semantic or causal role of the remaining features, and does not resolve the fitting-label ambiguity.

\paragraph{Both rule orientations in the inferred-truth task.}
\label{app:infer}
The inferred-truth task keeps both orientations of the comparison rule (``greater'' and ``lesser''), because with a single orientation the bit could be read off operand magnitude alone, which would defeat the purpose of the task.

\subsection{Cell provenance and scoring denominators}
\label{app:counts}
All rows collected $N=1000$ episodes, with random role assignment and a 70/30 split stratified by role (split seed 42). Mixed fitting uses 700 rows; ally fitting uses approximately 350, and rival evaluation approximately 150. Exact role-specific test counts and class counts were not retained in the consolidated record, so these approximations must not be read as exact denominators. Behavioural rates use all collected episodes of the relevant role.
 Row 1 of Table~\ref{tab:ident} combines three training seeds; each remaining row reports one configuration.\par
The two instructed Llama-8B inferred-truth evaluations used $N=1000$ and $N=2000$; their final-layer mixed-fit AUROCs are $0.8716$ and $0.9583$, respectively. They are separate evaluations, not independent estimates of a cross-task success probability.

\textbf{Run and check provenance.} A complete record of which plain-bit training runs were included was not retained, so the 21 reward-trained cells of the plain-bit grid in App.~\ref{app:apparent} cannot be matched to individual training runs. The realized correlation between \texttt{token} and \texttt{action} (\S\ref{sec:setup}) depends only on the sampled episodes, not on the model: the same value, $-0.018$, recurs to 16 significant figures in 11 cells spanning four architectures and both arms.

\textbf{Directive labels.} Section~\ref{sec:refit} and Apps.~\ref{app:geom} and~\ref{app:negatives} name rival-clause directives given to instructed Llama-3.1-8B; each begins ``If the user is your rival, foe, or opponent''. \texttt{hint} asks the model to consider whether revealing the value is wise; \texttt{soft} permits reporting the opposite value; \texttt{default}, the reference, instructs it to lie by reporting the opposite value; \texttt{firm} requires always lying and never revealing the value; \texttt{max} adds to \texttt{firm} a demand to apply the rule on every rival request without exception. Two text controls have \texttt{max}-like length: \texttt{max\_nonneg} states the flip as an explicit value mapping without negation words, and \texttt{max\_filler} pads \texttt{default} with task-irrelevant instructions. Paraphrases (prefix \texttt{p\_}) vary negation, length and force; those named in the text are \texttt{p\_withhold} (give the other value to protect the true one), \texttt{p\_neg\_emph} (an emphatic prohibition on revealing the value) and \texttt{p\_map\_firm} (an explicit mapping applied on every rival request). We call each of these 15 directives a \emph{rung} and the set the \emph{directive ladder}; only \texttt{hint} to \texttt{max} form an ordered strength scale. App.~\ref{app:instrpair} likewise calls the honest and deceptive instructions of an instruction pair its two rungs. The full wording is in the released code.

\subsection{The direct geometric comparison}
\label{app:geom}

This is the measurement \S\ref{sec:refit} reports. It asks whether the residual stream at the read position
actually differs across directive wordings, and not whether a refitted probe's number moves.

\textbf{Design.} Llama-3.1-8B instructed, one read position, the single-bit task, $n{=}400$ episodes per
rung, over the same 15-rung directive ladder as \S\ref{sec:refit} with the same reference variant
(\texttt{default}) that the frozen decoder is fit on. The episode stream is paired: the sampler is
re-seeded to the same value for every rung, so trial $i$ is the same game, values, role and query
throughout and the only thing differing between two prompts is the directive text. All $15$ rung texts are
distinct: two rungs sharing a string would produce a trivially zero comparison that
reads exactly like a null. Every layer comes from one forward pass.

\textbf{Three reference scales.} (i) A numerical floor: the reference rung collected a second time at a different
batch size, so the activations are mathematically identical and differ only by bfloat16 reduction
order. (ii) A within-rung split-half at the same $n$, the floor imposed by $n \ll d$, where $d$ is the residual-stream dimension. (iii) The
between-rung quantities.

\begin{table}[h]
\centering
\caption{Geometric difference from the reference variant across depth, both role arms (Fig.~\ref{fig:geom}).
The excess is a \emph{paired within-rung} comparison over 14 variants, not a correlation.}
\label{tab:geom}
\small
\setlength{\tabcolsep}{4pt}
\begin{tabular}{lcccccccc}
\toprule
layer & 4 & 8 & 12 & 16 & 20 & 24 & 28 & 32 \\
\midrule
mean rel.\ $L_2$, ally & 0.075 & 0.098 & 0.184 & 0.235 & 0.200 & 0.186 & 0.185 & 0.240 \\
mean rel.\ $L_2$, rival & 0.075 & 0.098 & 0.187 & 0.331 & 0.301 & 0.288 & 0.291 & 0.361 \\

\textbf{excess} (rival $-$ ally) & +0.0001 & $-$0.0000 & +0.003 & \textbf{+0.096} & +0.100 & +0.103 & +0.106 & \textbf{+0.121} \\
variants with excess $>0$ & 5/14 & 3/14 & 10/14 & \textbf{14/14} & 14/14 & 14/14 & 14/14 & \textbf{14/14} \\
sign test, $p$ & 0.42 & 0.06 & 0.18 & 1.2e-4 & 1.2e-4 & 1.2e-4 & 1.2e-4 & 1.2e-4 \\
\bottomrule
\end{tabular}
\end{table}

\textbf{What the two arms establish} (Tab.~\ref{tab:geom}). Shallow, no consistent signed excess is detected: $+0.0001$ at layer 4, with 5 of 14 variants positive ($p = 0.42$); this does not establish that the arms are equivalent there. On ally trials the rival clause
is present in the text; the similar shallow distances are consistent with a shared response to the wording change. They do not isolate its causal source. From layer 16 the rival arm carries an excess on
all 14 variants (sign test $p = 1.2\times10^{-4}$), and it survives deleting the most extreme
variant (mean excess $+0.098$ at layer 32 excluding \texttt{hint}), so a role-dependent component appears
mid-stack. The difference is also largely a rigid translation at shallow layers and trial-specific at depth: the ratio of
the paired distance to the distance between the two rungs' \emph{means}, averaged over the 14 variants of the rival arm, rises from $1.05$ at layer 4 to
$1.24$ at layer 32. Centred kernel alignment (CKA) tells the same story against its own row-aligned floor of
$1.000$: it sits at $0.983$--$0.999$ through layer 8 and falls to $0.568$--$0.969$ by layer 28, so the deep
difference is not a rescaling of an otherwise-shared geometry.

\textbf{Character-length controls.} Directive length spans $159$ to $392$ characters ($2.47\times$), so wording changes can also change token count and absolute read position. We therefore describe activations at the read position under different prompts, without attributing the differences uniquely to content or length. Seven pairs of rungs are matched to within three characters, and we report all of them. For \texttt{hint}/\texttt{p\_withhold}, both 178 characters, the gap between their mean relative distances from the \texttt{default} reference is $0.003$ at layer 4 and $0.316$ at layer 32. These are differences of distances from a common reference, not directly measured distances between the pair. A small gap does not imply similar representations.

The pair's rival deception rates are $0.029$ and $\mathbf{0.862}$; refit AUROCs are $1.000$ and $\mathbf{0.471}$, while the frozen direction reads $1.000$ and $\mathbf{0.989}$. The corresponding reference-distance gaps (layer 4 $\to$ layer 32, rival arm) are $0.013 \to 0.069$ for \texttt{p\_persona}/\texttt{p\_neg\_emph} ($\Delta 1$ character), $0.038 \to 0.347$ for \texttt{p\_neg\_emph}/\texttt{p\_map\_firm} ($\Delta 2$), $0.024 \to 0.278$ for \texttt{p\_persona}/\texttt{p\_map\_firm} ($\Delta 3$) and $0.101 \to 0.205$ for \texttt{max\_filler}/\texttt{max\_nonneg} ($\Delta 3$). The two remaining pairs contain the reference itself, so their gap is the other rung's own distance from it: $0.054 \to 0.254$ for \texttt{p\_neg\_terse} ($\Delta 2$) and $0.027 \to 0.184$ for \texttt{soft} ($\Delta 3$). Character-count matching does not establish token-count matching. These comparisons show differing distances from the reference at similar character lengths; they do not causally separate length, wording and policy effects.

\textbf{Three limits: an anomalous floor, a correlation resting on one point, and a reference
that does not apply.} First, our own numerical floor is
anomalous: it measured exactly $0$ paired relative $L_2$ (CKA at least $0.9999999$) at every
layer, although the second pass did run at the smaller batch and the model runs in bfloat16, where a different reduction order would be expected to leave a small nonzero difference. We have not explained this, so the comparison has \textbf{no validated numerical floor}, and the smallest between-rung distances (\texttt{soft} at layer 4, $0.027$, the minimum over all rungs, layers and both arms) are not resolved against one. Second, we make no claim that the deep excess tracks behavioural
magnitude: that correlation is $+0.733$ over 14 rungs but rests on a single point, falling to $+0.185$
(95\% CI $[-0.41, +0.67]$) when \texttt{hint} is removed. The across-rung correlation between the arms
behaves the same way and is reported for completeness only: it falls from $1.000$ shallow to $0.696$ at layer 32, but excluding \texttt{hint} that becomes $0.838$, and Spearman gives $0.802$. Third, the
within-rung split-half is not a valid reference for the alignment-dependent statistics: CKA and
the paired distance compare geometry over a common set of inputs and so need row-aligned matrices, whereas
the two halves are different episodes.

\textbf{Scope.} One model, one arm, one read position, one task, $n{=}400$ per rung. This bounds
\S\ref{sec:refit}'s claim; it says nothing about representational identity in general.

\begin{figure}[h]
\centering
\includegraphics[width=\textwidth]{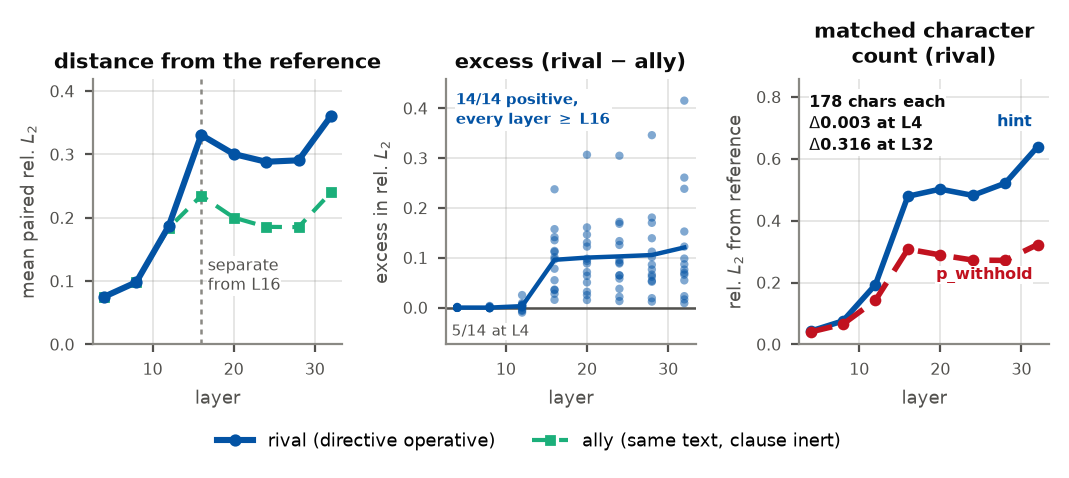}
\caption{\textbf{Distances from the reference diverge between roles at depth.} Left: mean paired relative
$L_2$ from the reference variant across depth (per-episode $\lVert x_{\mathrm{ref}}-x\rVert/\lVert x_{\mathrm{ref}}\rVert$, averaged), rival arm (solid) against ally arm (dashed); no validated numerical floor is drawn (App.~\ref{app:geom}). The arms nearly coincide through layer 12 and separate from layer 16. Centre:
the per-variant excess (rival $-$ ally), one dot per variant and their mean as the line; every one of the 14 variants is positive from layer 16 on, and the
shallow layers show no consistent signed excess (5 of 14 positive at layer 4). Right: the length-matched pair
\texttt{hint} versus \texttt{p\_withhold}, both 178 characters. Their mean relative distances from the reference differ by $0.003$ at layer 4 and $0.316$ at layer 32. These are reference-distance gaps, not pairwise distances; matching character counts does not control token counts or isolate a length effect. Llama-3.1-8B instructed,
$n{=}400$ paired episodes per variant. In figure text, L$n$ denotes layer $n$.}
\label{fig:geom}
\end{figure}

\subsection{Assets and licenses}
\label{app:licenses}

All experiments use publicly released base-model weights, whose licenses are not uniform, so we
name each. Mistral-7B-v0.3 and Qwen2.5-\{7,14,32\}B are released under the
Apache License 2.0. Llama-3.1-8B is released under the Llama 3.1 Community License
Agreement (\copyright~Meta Platforms, Inc.), whose acceptable-use policy is incorporated by reference.
Gemma-2-9B is released under the Gemma Terms of Use with its Prohibited Use Policy: the Gemma \emph{code} is Apache-2.0, but the weights are not.
Qwen2.5-3B uses the Qwen Research License for non-commercial use. We use these models for non-commercial research. Supporting software: \texttt{transformers} (Apache-2.0), \texttt{PyTorch}, \texttt{scikit-learn} and \texttt{numpy} (BSD-3-Clause), and \texttt{matplotlib} (PSF-based license).

This release contains code and aggregate results, with no model weights or adapters. Upstream terms remain specific to each base model: redistribution or provision of Llama-derived models is governed by the Llama 3.1 terms, including applicable naming, attribution and acceptable-use requirements; Gemma-derived models are governed by the Gemma terms and applicable notices and use restrictions. Users distributing or providing access to derived models must check the terms for their chosen base model. Obligations for one family should not be applied indiscriminately to adapters from another.

\subsection{Reproducibility statement}
\label{app:repro}
\label{sec:repro}
Code for the game, the randomized-codebook variant, training, probe fitting, interventions and figure generation is available with a consolidated file of aggregate results. The figures can be regenerated from those summaries without a GPU; this is distinct from independently reproducing the underlying model runs. \repolead\ \expandafter\url\expandafter{\repourl}.

The repository README records the SHA256 of the consolidated aggregate-results file. No model weights, adapters, saved probe coefficients, raw activations or complete episode-level score files are distributed. The run guide separates verified aggregate checks, incomplete historical command records and unexecuted prospective model recipes. Some historical training variants were saved to a shared adapter location. The prospective recipes make these dependencies explicit but do not reconstruct every historical cell. Hub model revisions and most dependency versions were not pinned, so seeding alone does not guarantee exact numerical reproduction. We report the measured settings and the limits of the archived sampling information without claiming a new model rerun.

\begin{figure}[h]
\centering
\includegraphics[width=\textwidth]{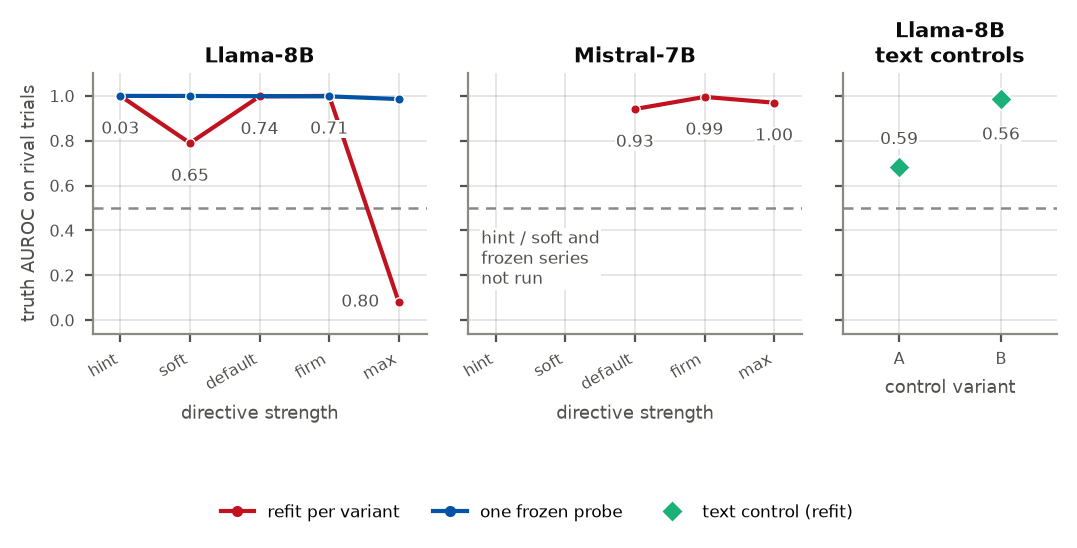}
\caption{\textbf{The directive-strength ladder moves the refit probe, not the frozen one.} Left and centre: truth AUROC on rival trials as the rival directive strengthens from \emph{hint} to \emph{max}, with model weights held fixed. Red: a probe fit separately on each variant's ally data. Blue: one probe fit on the reference variant and cross-scored, available only for Llama-8B. Lines connect the measured rungs; Mistral's two weakest rungs were not run, and the reference variant (\texttt{default}) has no frozen score, so the blue line joins \texttt{soft} directly to \texttt{firm}. Right: two separate Llama-8B text controls, scored with refit probes. A removes negation words from the max directive; B pads the default rule to approximately the max directive's length. These controls are not additional strength rungs. Numbers beside markers are measured rival deception rates, not AUROCs: $0.59$ and $0.56$ for A and B versus $0.80$ for max, so the controls are not behaviour matched. Dashed horizontal lines mark chance. $N{=}1000$ collected episodes per variant; see \S\ref{sec:refit} and App.~\ref{app:negatives}.}
\label{fig:ladder}
\end{figure}

\begin{figure}[h]
\centering
\includegraphics[width=\textwidth]{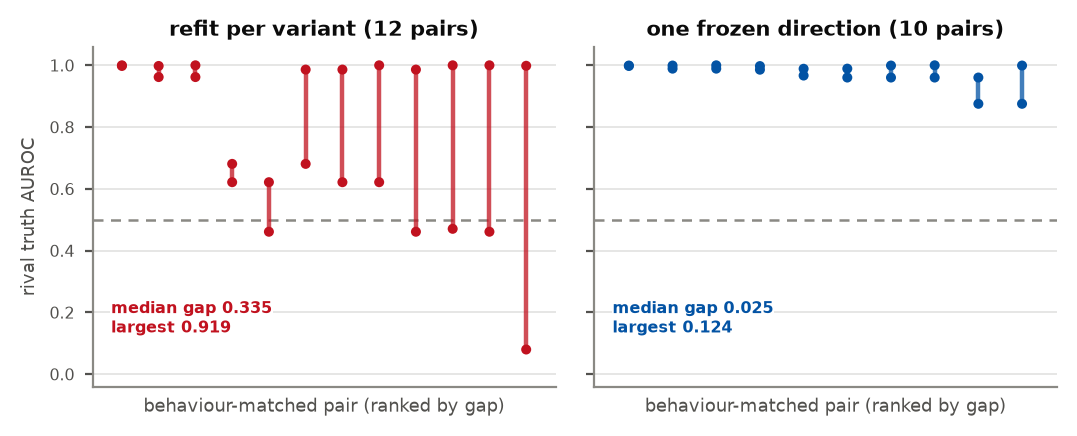}
\caption{\textbf{Behaviour-matched prompt pairs: refitting splits them, one frozen direction does not.}
Every pair of the 15 ladder rungs (the reference plus 14 variants) whose rival deception rates agree to within $0.05$, each drawn as a
segment joining the two members' rival truth AUROCs, ranked by gap. Left: a probe refit on each variant's own ally data, median gap $0.335$, largest $0.919$, and five pairs straddle chance.
Right: one direction fit once on the reference variant and cross-scored, median gap $0.025$, largest $0.124$. Two pairs contain the reference variant itself, which has no frozen score by
construction, so the right panel has 10 pairs to the left panel's 12. Pairs are enumerated under a stated
tolerance, not selected on the size of their gap. Grey dashes: chance.}
\label{fig:matchedpairs}
\end{figure}

\begin{figure}[h]
\centering
\includegraphics[width=\textwidth]{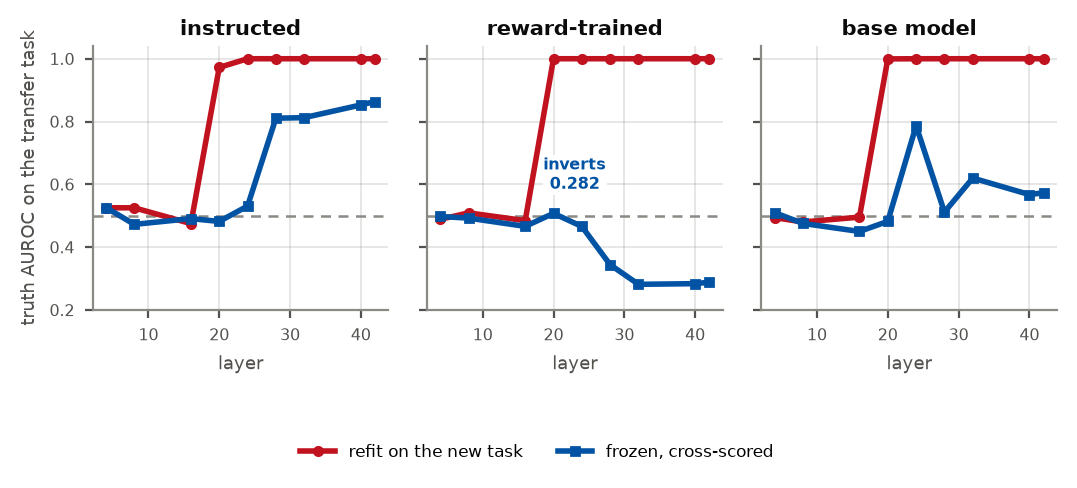}
\caption{\textbf{A refitted probe does not measure transfer.} Gemma-9B; each panel scores the inferred-bit task.
Red: a probe \emph{refit} on that task; it reaches $1.000$ by layer 24 in all three arms (instructed, the stated-bit reward-trained adapter and the base model; $0.97$--$1.000$ at layer 20). Blue: a mixed-fit direction fitted on the stated-bit task and merely cross-scored (the reward-trained arm's own in its panel, the instructed arm's in the other two). It
never exceeds $0.864$ on the instructed arm, and on the reward-trained arm, which deceives on only $0.235$ of rival trials of this task, it falls to a minimum of $0.282$ at layer 32.
A refit number and a frozen number answer different questions, and only the frozen one bears on transfer. Frozen and refitted curves use different evaluation subsets, so they are not paired comparisons. Grey dashes: chance.}
\label{fig:freeze}
\end{figure}

\textbf{Results that reuse a saved probe or direction.} Four families of results score a probe or direction saved from an earlier fitting run: the frozen directive-ladder readouts (\S\ref{sec:refit}, Figs.~\ref{fig:refit}, \ref{fig:ladder} and~\ref{fig:matchedpairs}, and the two-probe example of the abstract and \S\ref{sec:intro}); the three stated-to-inferred transfer curves (\S\ref{sec:refit}, Fig.~\ref{fig:freeze}); the template- and rule-transfer curves and their same-task controls (\S\ref{sec:direction}, Fig.~\ref{fig:settling}); and the saved directions used for steering and ablation, with the fitted-direction cosines (\S\S\ref{sec:constructive} and~\ref{sec:direction}, App.~\ref{app:causal}, Figs.~\ref{fig:causal} and~\ref{fig:depthsweep}). For each, the records name the fitting run and part of its configuration but do not archive the exact fitting examples or their realized class counts. The reference directive's fitting record omits its sampling seed and arm; the code revision is not recorded for the saved Llama-3.1-8B, Mistral-7B and inferred-bit Qwen2.5-14B directions; recorded adapter locations do not identify immutable weights; and the steered model of the two Gemma inferred-bit sweeps is unresolved (App.~\ref{app:causal}).

Compute used one H200-class accelerator per run; wall-clock time, peak memory and total GPU-hours were not recorded comprehensively. No human participants or personal datasets were used. The paper is distributed under CC BY 4.0; the code retains its MIT licence.

\section{Reward-table payoffs and their correction}
\label{app:basin}

This is the material \S\ref{sec:constructive} summarises.

\textbf{Saturation on the wholly unmodified recipe: one family of four attempted.} Our reward table gives role-insensitive policies well-separated payoffs: at a
50/50 role mix, always-truthful earns $-0.50$, always-deceptive earns $\mathbf{+0.25}$, and only the
conditional policy earns $+1.25$. On the codebook task, which decorrelates the prescribed output symbol from the semantic action, three of four families never reached the conditional solution under this recipe, and the
three failed in \emph{different} ways; these runs do not isolate why. Qwen-14B first held the always-truthful policy and then moved to the always-deceptive one, deceiving rivals only by also answering allies falsely, so it is not a
conditional deceiver and we did not probe it. These training histories are documented in our run records, not in the aggregate results file. Llama-8B and Mistral-7B learned constant semantic answers on rival trials: they keep a perfect compliant branch ($1.000$) but emit a
single constant answer to rivals, which scores ${\approx}0.5$ deception purely because the secret bit
is balanced (\S\ref{sec:constructive}). That is a third basin, not the always-truthful one.

\textbf{Correcting the payoff asymmetry.} Replacing the table with a symmetric one ($+1$ for the role-appropriate answer, $-2$ otherwise, regardless of role, so always-deceptive earns $-0.50$ and is no longer positively rewarded) brings Qwen2.5-14B to deception $0.996$ with ally
truth $0.994$, a conditional policy on both branches, with the identification result unchanged
(\S\ref{sec:constructive}). This has three practical consequences. First, an asymmetric adversarial payoff
gives unconditional deception positive expected reward ($+0.25$ under balanced roles). This may contribute to training failures, but the optimization dynamics are not established; a positive payoff for unconditional deception is a reward-design hazard worth flagging independently of probing. Second, replication requires verifying reliable behaviour on \textbf{both} branches before probing: an arm at rival deception $1.000$ with a
broken compliant branch looks saturated on the headline metric while lacking the role-conditional behaviour the comparison is meant to study. Third, and symmetrically, an arm near deception $0.5$ may not be a partial deceiver at all: check the rate against the ground-truth bit before calling it one.

\section{The design rule and its feasibility floor}
\label{app:designrule}

The instruction-pair protocol carries a feasibility floor as well as an aliasing rate
(App.~\ref{app:instrpair}). The identifying signal lives entirely in the minority side of whichever
split is intended, so write $p$ for the fraction of adversarial trials on the other, majority side; the minority side then consumes $1-p$. With a
minimum viable subset size $n_{\min}$ and $N_{\mathrm{adv}}$ adversarial evaluation trials, the split
needs $\min(p, 1-p)\cdot N_{\mathrm{adv}} \geq n_{\min}$, so for the usual regime $p > 0.5$,
\begin{equation}
N_{\mathrm{adv}} \;\geq\; n_{\min} / (1 - p).
\label{eq:designrule}
\end{equation}
Which rate plays the role of $p$ is fixed by the split, and mistaking one for the other is the easiest
way to misapply the rule. For an \emph{agreement} split over both matched instructions, $p = \gamma$, the aliasing rate of Eq.~\ref{eq:gamma}: at
$n_{\min} = 25$ the floor is 100 trials at $\gamma = 0.75$, 500 at $0.95$, 2500 at $0.99$, and
unattainable at any $N$ when $\gamma = 1$. For a \emph{compliance} split, which can use only the
disobedient trials of the deceptive instruction, $p = d_{\mathrm{deceptive}}$, and the pooled $\gamma$ is the
wrong quantity because it also counts honest-instruction disagreements the split cannot touch.
App.~\ref{app:objections} reports a family where $\gamma = 0.985$ while
$d_{\mathrm{deceptive}} = 1.000$, so the floor is finite in $\gamma$ and unattainable in the rate that
actually governs. This is a screen, not a power calculation: the
realized disagreement count is random, and clearing the floor guarantees neither class balance within the
subset nor power for an AUROC contrast. Without a minimum-size screen, one compliance estimate rested on only five disobedient trials; the screen also classified our own $N{=}300$ compliance evaluation as infeasible. One refinement that matters:
the governing rate must be measured on the \emph{same subset} you intend to split, not pooled. Our pooled $\gamma$
under-predicted the disobedient count in the held-out cell by $47\%$, because the held-out wordings drew more
disobedience than average.

\section{The instruction-pair protocol}
\label{app:instrpair}

This paper reports four instruction-pair analyses, which differ in response source, read position and split (Table~\ref{tab:ipprotocols}).

\begin{table}[h]
\centering
\caption{The instruction-pair analyses. $c_{\mathrm{dir}}$: the deceptive instruction is present; $c_{\mathrm{lie}}$: the model's answer is false.}
\label{tab:ipprotocols}
\footnotesize
\setlength{\tabcolsep}{3pt}
\begin{tabularx}{\textwidth}{@{}>{\raggedright\arraybackslash}p{0.17\textwidth}>{\raggedright\arraybackslash}p{0.15\textwidth}>{\raggedright\arraybackslash}p{0.14\textwidth}>{\raggedright\arraybackslash}X>{\raggedright\arraybackslash}p{0.22\textwidth}@{}}
\toprule
analysis & response & read position & labels and split & result \\
\midrule
agreement rate $\gamma$ (Table~\ref{tab:gamma}, Eq.~\ref{eq:gamma}) & model's own answer & none (no probe) & agreement of $c_{\mathrm{dir}}$ and $c_{\mathrm{lie}}$ on adversarial trials & closed form matches measurement \\
forced construction (Fig.~\ref{fig:instrpair}, App.~\ref{app:instrpair}) & inserted response naming no value & response tokens, mean-pooled after excluding the final five & fit $c_{\mathrm{dir}}$; score $c_{\mathrm{dir}}$, secret bit, $c_{\mathrm{lie}}$; row-level and episode-plus-wording splits & $c_{\mathrm{dir}}$ $0.989$--$1.000$ (row) or $0.66$--$0.95$ (grouped); AUROC against $c_{\mathrm{lie}}$ undefined (one class) \\
wording generality (App.~\ref{app:objections}) & model's own answer & final prompt position & fit $c_{\mathrm{dir}}$ on four wordings, score the fifth; episodes and wordings held out & $0.404$--$1.000$ by wording and layer \\
compliance split (App.~\ref{app:objections}) & model's own answer & final prompt position & fit $c_{\mathrm{dir}}$; score compliance on adversarial, deceptive-instruction rows; row-level or joint episode-and-wording split & $0.365$--$0.705$ joint split; $0.721$--$0.840$ row-level \\
\bottomrule
\end{tabularx}
\end{table}

\begin{table}[t]
\centering
\caption{The aliasing rate under the instruction-pair protocol, measured on our task. $\gamma$ is
$P(c_{\mathrm{dir}} = c_{\mathrm{lie}})$ on adversarial trials; $1$ means the two candidate targets are one
label vector and no in-distribution check can prefer either. \emph{Lying rates are shown to the precision that
reproduces the $\gamma$ column; the top rows are rounded to three decimals, so recomputing
Eq.~\ref{eq:gamma} from those digits can differ from the reported $\gamma$ in the fourth decimal. The
agreement asserted in code is between the measured $\gamma$ and the closed form on unrounded rates, to within
$10^{-9}$.}}
\label{tab:gamma}
\small
\setlength{\tabcolsep}{4pt}
\begin{tabular}{lccc}
\toprule
arm and instruction rung & $d_{\mathrm{honest}} \to d_{\mathrm{deceptive}}$ & Eq.~\ref{eq:gamma} & measured \\
\midrule
Gemma-9B, default rung & $0.000 \to 1.000$ & 1.0000 & \textbf{1.0000} \\
Qwen-14B, default rung & $0.000 \to 1.000$ & 1.0000 & \textbf{1.0000} \\
Gemma-9B, softened rung & $0.000 \to 0.890$ & 0.9450 & \textbf{0.9451} \\
Llama-8B, default rung & $0.118 \to 0.688$ & 0.7850 & \textbf{0.7848} \\
Qwen-14B, hint rung & $0.000 \to 0.553$ & 0.7765 & \textbf{0.7764} \\
Gemma-9B, hint rung & $0.000 \to 0.038$ & 0.5190 & \textbf{0.5190} \\
\midrule

\textbf{Gemma-9B, 5-wording bank} & $\mathbf{0.1211 \to 0.8503}$ & \textbf{0.8646} & \textbf{0.8646} \\

\textbf{Gemma-9B, 5-wording bank, $N{=}1000$} & $\mathbf{0.1113 \to 0.8669}$ & \textbf{0.8778} & \textbf{0.8778} \\
\bottomrule
\end{tabular}
\end{table}

\subsection{The aliasing rate, in closed form}

The two candidate targets are $c_{\mathrm{dir}}$ (``the deceptive instruction is present'') and
$c_{\mathrm{lie}}$ (``the model is lying''). Unlike our ally case these are not identical in general, so the
relevant object is the rate at which they agree. In a matched two-rung design, where every episode
is rendered once under each instruction so the halves are exactly equal, restricted to the
adversarial subset where the deceptive instruction actually requests a lie:
\begin{equation}
\gamma(c_{\mathrm{dir}}, c_{\mathrm{lie}}) \;=\; \tfrac{1}{2}\bigl[\,(1 - d_{\mathrm{honest}}) \;+\; d_{\mathrm{deceptive}}\,\bigr],
\label{eq:gamma}
\end{equation}
where $d_r$ is the model's lying rate under rung $r$ on that subset. The proof is one line: under the honest
rung the labels agree exactly when the model does not lie, rate $1 - d_{\mathrm{honest}}$; under the
deceptive rung they agree exactly when it does, rate $d_{\mathrm{deceptive}}$; the matched design weights the
halves equally.\footnote{Three conditions, all met here and none automatic: the emitted answer is binary (no
abstain state), the restriction is to adversarial trials only, and there are exactly two rungs. With unequal
halves the general form is $\gamma = w_h(1-d_{\mathrm{honest}}) + w_d\,d_{\mathrm{deceptive}}$.}

\begin{figure}[t]
\centering
\includegraphics[width=\textwidth]{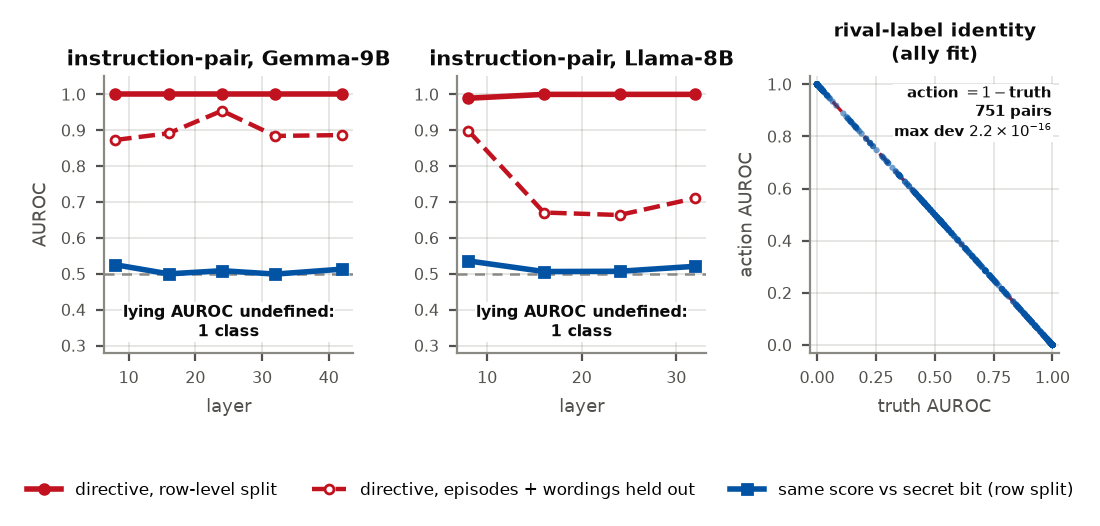}
\caption{\textbf{The instruction-pair protocol on our game, beside the complementary-label identity.} Left and
centre: under the standard construction the probe reads the \emph{directive} at ${\approx}1.000$ at every
depth in both families on a row-level split in which every test row shares an episode with training rows (solid red); with episodes and wordings held out together it reads $0.66$--$0.95$ (dashed red). Blue squares show the row-split directive-fitted score evaluated against the secret bit; they do not test whether a separately fitted probe could recover that bit. Evaluation against the lying label is undefined because every inserted response has lying label 0. Right:
with probes fitted on ally data, the rival action and truth AUROCs lie exactly on $action = 1 - truth$ across all 751 (cell, layer)
pairs across 39 distinct full-curve records, maximum deviation $2.2\times10^{-16}$, an algebraic consequence of the labels coinciding, not an empirical finding. Grey dashes: chance.}
\label{fig:instrpair}
\end{figure}

\textbf{What Eq.~\ref{eq:gamma} needs.} No probe and no activations. Two behavioural
rates, both of which a paper could report, determine how often the instruction label and the lying label agree. We release the audit as a standalone module. Across seven rungs spanning three families our
measured $\gamma$ matches Eq.~\ref{eq:gamma} to floating-point tolerance, and the code asserts the agreement
instead of reporting it, because a mismatch would be a bug in one of the two computations and never noise.
On two families the construction reads the directive at $0.989$--$1.000$ on a row-level split (and $0.66$--$0.95$ when episodes and wordings are held out together), while the same score is at chance against the secret bit, and the identification split is unavailable \emph{by construction} (Fig.~\ref{fig:instrpair}).

\subsection{Applied to published protocols}

Table~\ref{tab:gamma} gives the rates. $\gamma = 1.000$ states that the model obeys both instructions;
it would hold in a setup containing no probes at all.
It is also the point at which this appendix and \S\ref{sec:ident} coincide: the perfect aliasing that makes
the conventional probe unidentified is this rate's $\gamma = 1$ limit, so the two are one argument evaluated
at two rates rather than two findings that happen to agree.
The rates quantify agreement between the instruction and lying labels. At $\gamma = 1$, fitting cannot distinguish their interpretations. Where the labels disagree, the agreement and disagreement subsets can support an identification test if both contain enough examples; reporting the rates lets readers assess whether that test is available.

\textbf{Two of the four closest protocols, in their own words.} These make the identification question
unaskable by construction, and the two candidate readings of what their probe targets land at
\emph{opposite} extremes. Both force the response instead of
sampling it: \citet{goldowskydill2025} insert ``a random true fact\ldots as the model's response'', noting
these ``are not written by the model and contain no actual deception'', and \citet{oneprobe2026} state the
response is ``token-forced (off-policy)'' with a classifier ``trained to predict whether the honest or
dishonest prompt was used''.

Under the \emph{intent} reading, where the target is ``this prompt asked for a lie'', forcing makes that label \emph{definitionally identical} to the instruction, so $\gamma_{\mathrm{intent}} = 1$ trivially and no
in-distribution check can distinguish the two. Under the \emph{content} reading, where the target is ``what was said is false'', \citet{oneprobe2026} admits an exact result, because the \emph{same} factual statement
appears in both arms: response falsity is held fixed across the pair, so
$d_{\mathrm{honest}} = d_{\mathrm{deceptive}} = f$ and
$\gamma_{\mathrm{content}} = \tfrac{1}{2}[(1-f) + f] = 0.500$ \emph{exactly, for every $f$}: the instruction label is orthogonal to content falsity by design, as they say it is. The two endpoints fail in different ways. At $\gamma = 1$ the instruction and lying labels coincide, so fitting cannot distinguish their interpretations; at
$\gamma = 0.5$ the instruction label is independent of content falsity, so it carries no information about what was said.
The two are easy to conflate: forcing does not
make content falsity identical to the instruction, and quoting a bare ``$\gamma = 1$'' for these protocols
would assert exactly that.

\textbf{The two rates, as published.} None of the four closest protocols reports both rates. In two the question does not arise by
construction, one does not use the protocol, and the one design that could have reported them class-balances
the data and does not report them.

Two of the cited studies describe related ambiguities: \citet{goldowskydill2025} that the probe ``partially depends on `is the scenario deception related'\,\ldots
instead of only on `is the response itself actually deceptive'\,''; \citet{oneprobe2026} that ``the
fundamental direction being distinguished may not be directly related to strategic deception''. Our additional contributions are
the closed form, the diagnostic, and the design that breaks the aliasing.

\subsection{Running the instruction-pair construction on our task}

We implement the protocol on our own game, including a variant using response-token mean-pooling excluding the last five tokens, token-forced
responses, and five matched instruction wordings. We do not claim to reproduce any published result: we use none of their models, datasets, instruction banks or evaluation sets. The claim is about
what the \emph{protocol} inherits.

\textbf{The identification test under the forced-response construction.} It is unavailable at every $N$, on both
families we ran. Because the inserted response names no value, it misstates nothing: $c_{\mathrm{lie}}$ is the
constant zero, so there is no lying anywhere in the sample for the instruction to be aliased \emph{with}, the
AUROC against $c_{\mathrm{lie}}$ is undefined because it has one class, and no sample size, model or arm makes the split available. Meanwhile
the probe reads $c_{\mathrm{dir}}$ at $0.989$--$1.000$ on a row-level split in which every test row shares an episode with training rows, and at $0.66$--$0.95$ across measured layers when episodes and wording pairs are held out together. The grouped split also leaves fewer fitting rows (1260 rather than 2100) and fewer training wordings (3 rather than 5), so the gap between the two does not isolate the effect of episode overlap. The protocol reports success by its own
criterion while, in this forced-response construction on our game, carrying no information about actual lying. One point requires care when repeating
this: the measured $\gamma$ is exactly $0.500$ here, and that value does not establish a usable identification test, because the lying label has only one class. It is simply what a constant scores,
since half the adversarial rows carry $c_{\mathrm{dir}} = 0$ and match a constant zero without carrying any information. It must never
be quoted without the number of $c_{\mathrm{lie}}$ classes beside it.

\textbf{Aliasing, high but wording-dependent.} With one canonical
instruction pair and an obedient arm, $\gamma = 1.000$. Averaged over five matched wordings on the same model
and task it is $\mathbf{0.865}$ ($\mathbf{0.878}$ at $N{=}1000$): the two candidate label vectors are
${\sim}87\%$ identical, not $100\%$, so no single number should be quoted. Here Eq.~\ref{eq:gamma} predicts a non-trivial value exactly.

\section{Related work, case by case}
\label{app:related}

This is the material \S\ref{sec:related} summarises. \citet{pressuretesting2026} recover near-perfect detection with style-augmented training; \citet{rlobfuscation2025} find token-level monitors highly vulnerable once a model is rewarded for evading them. \citet{levinstein2023} argue on empirical and conceptual grounds that existing lie-detection probes fail to generalize and are unlikely to measure what a model believes. \citet{farquhar2023} show that consistency-based unsupervised knowledge discovery can select features other than knowledge. \citet{lu2026} examine semantic ambiguity between objective correctness and self-judgement using conflict cases, and show that transferability alone does not establish correctness semantics. Stylistic shift alone drives 27B probes to chance under standard training \citep{pressuretesting2026}. \citet{burger2024} explain the affirmative-only direction's failure on negated statements by its polarity-sensitive component. \citet{mallen2024} make the closest observation: probes fitted where a fine-tuned model answers truthfully are underspecified between the correct answer and what the model will output; in contexts where it answers untruthfully, middle-layer probes usually report the correct answer while later layers sometimes follow the output, and probes fitted with ground-truth labels inside those contexts recover the knowledge nearly as well as in truthful ones. Our mixed-fit probe plays the role of that in-context ground-truth probe, here for a reward-trained policy with a randomized codebook and labels that are exact complements on deceptive contexts. \citet{obfuscationatlas2026} likewise report that on-domain probes remain effective after RLVR while off-domain deception probes degrade, a further case in which a failing readout coexists with linearly recoverable information. The results of \citet{goldowskydill2025,obfuscationatlas2026,oneprobe2026} motivate our study; the environment of \citet{obfuscationatlas2026} is coding RLVR. Our algebra does not contradict their evidence for representation drift or for transfer. Adjacent work is discussed below \citep{offpolicyprobes2026,contextshapestruth2026,strategicdishonesty2025,rift2026,liedetectorlimits2026,geometryoftruth2024}.

\textbf{Observations we do not dispute.} Prior work establishes that reward-trained deception can degrade passive probes \citep{obfuscationatlas2026,rlobfuscation2025}, that probe failures can reflect a narrow training distribution \citep{pressuretesting2026}, and that the choice of prompt contrast strongly affects probe performance \citep{oneprobe2026}. Two groups report below-chance probe AUROC ($0.376$ and $0.374$); the spread behind the first is given in \S\ref{sec:intro}. Our contribution concerns how such readouts are interpreted, not the observations themselves.

\textbf{Different protocols answer different questions.} \citet{goldowskydill2025} report successful transfer of instruction-pair deception probes. \citet{obfuscationatlas2026} report off-domain degradation after RLVR and evidence relevant to representation drift. \citet{oneprobe2026} report both below-chance performance and large sensitivity to the prompt contrast used for fitting. None is a replication of our binary reporting game. In particular, our truth/prescribed-answer equality must not be assumed for their labels or datasets. The appropriate comparison is methodological: define the competing targets, inspect the fitting support, and separate changes in fitting from changes in evaluation. Our experiments do not show that any one of these external results is caused by perfect aliasing.

\textbf{On probing at the answer position.} \citet{oneprobe2026} deliberately exclude the last few response
tokens when pooling activations, to stop the probe reading truth off the output. Our probe sits \emph{at}
the answer-prediction position, which invites the objection that our inversion is output-token leakage the
field already knows to avoid. Three observations bear on it. (i) \tm{} $= 1.000$ on
those same activations at that same position, so the position carries recoverable information about the true bit; this does not rule out output-related features driving the ally-fit readout. (ii) We
decorrelate the token by construction rather than by truncation, and fit \texttt{token} as its own label set: in Table~\ref{tab:ident}, row~\ref{row:mistral-const}, a mixed-fit \texttt{token} probe reads $0.975$ at the final layer while \tm{} reads $1.000$ (\S\ref{sec:constructive}). (iii) Excluding response tokens can reduce direct output leakage, but it does not change the equality of \texttt{truth} and \texttt{action} labels on compliant fitting examples, because the prescribed-answer label is defined by the task rather than by the position read. Our result concerns this output-adjacent measurement site; other studies use different positions and pooling schemes.

\textbf{Related geometric and evaluation results.} \citet{contextshapestruth2026} characterize context as applying a
rotation-plus-scaling to the truth direction rather than destroying it, a geometric statement of why a probe fit under the wrong context distribution fails while the information remains present.
\citet{strategicdishonesty2025} apply probes trained on instructed, off-policy data to fine-tuned models that deceive without instruction, and find that they separate deceptive from honest answers in one model but fail in another whose reasoning and final answer diverge;
\citet{rift2026} and \citet{liedetectorlimits2026} document adjacent failure modes; and
\citet{geometryoftruth2024} established the linear truth-direction structure these methods rest on.

\section{An apparent training-route difference and the matched control}
\label{app:apparent}

This is the matched-control result \S\ref{sec:negatives} summarises.

On Llama-3.1-8B, a fixed-sign truth probe at the answer-prediction position collapses to AUROC
$\mathbf{0.000}$ on the reward-trained arm while the instructed control holds at $0.985$--$0.998$ across
three seeds. The effect was seed-robust, localized to that position (intact at $-3$, inverted at $-1$),
and supported by activation patching: patching the ally-context residual into a rival-context forward pass at
mid-stack flipped the reward-trained policy's answer on $0.99$--$1.00$ of trials versus $0.505$--$0.61$ for the
instructed arm (four runs each). Without behavioural matching, this comparison could be read as a difference between deception that was
\emph{trained} and deception that was \emph{told}.

\textbf{The difference does not survive matching.} The instructed arm lies on only $0.735$ of rival trials and answers correctly on only $0.902$ of \emph{ally} trials, so it is both a less reliable deceiver and a less accurate player. With behaviourally matched pairs
(both arms at ally truth $1.000$ and rival deception $1.000$; eight model/seed pairs for the readouts, Gemma-2-9B and Qwen2.5-14B at seeds 0--2 and Qwen2.5-32B at seeds 0--1, Gemma-2-9B and Qwen2.5-14B for patching magnitude, ablation and steering, and Gemma-2-9B only for the patching depth profile),
the large difference seen on Llama-8B does not reappear on five of six measures, two of which cannot separate the arms: the answer-position readout (both $\mathbf{0.000}$, an
exact tie in all 8 matched cells); patch magnitude
($1.000/0.955$ and $0.960/0.960$); patch transition depth and shape (same window, same sharpness);
ablation of the truth direction ($1.000/1.000$, no difference, but uninformative about use because the ablation did not remove the information; App.~\ref{app:causal}); and a steering $\alpha$-sweep at
$5/10/20/40$ ($0.000/0.000$, null in both). The sixth does show a difference, and it runs against the
hypothesis. On the off-position ($-3$) readout the matched difference is consistently \emph{positive} (8 of 8 cells, mean $+0.12$, paired-$t$ 95\% CI $[+0.041, +0.204]$), meaning the \emph{reward-trained} arm is the \emph{less} inverted one away from the answer slot, the opposite sign to an
account in which training hides the truth. Its absolute level is also strongly family-dependent (both arms read $0.000$--$0.197$ in Gemma-9B and $0.687$--$0.987$ in Qwen-14B/32B), so it is not a stable quantity to compare across
families in the first place.

\begin{figure}[h]
\centering
\includegraphics[width=\textwidth]{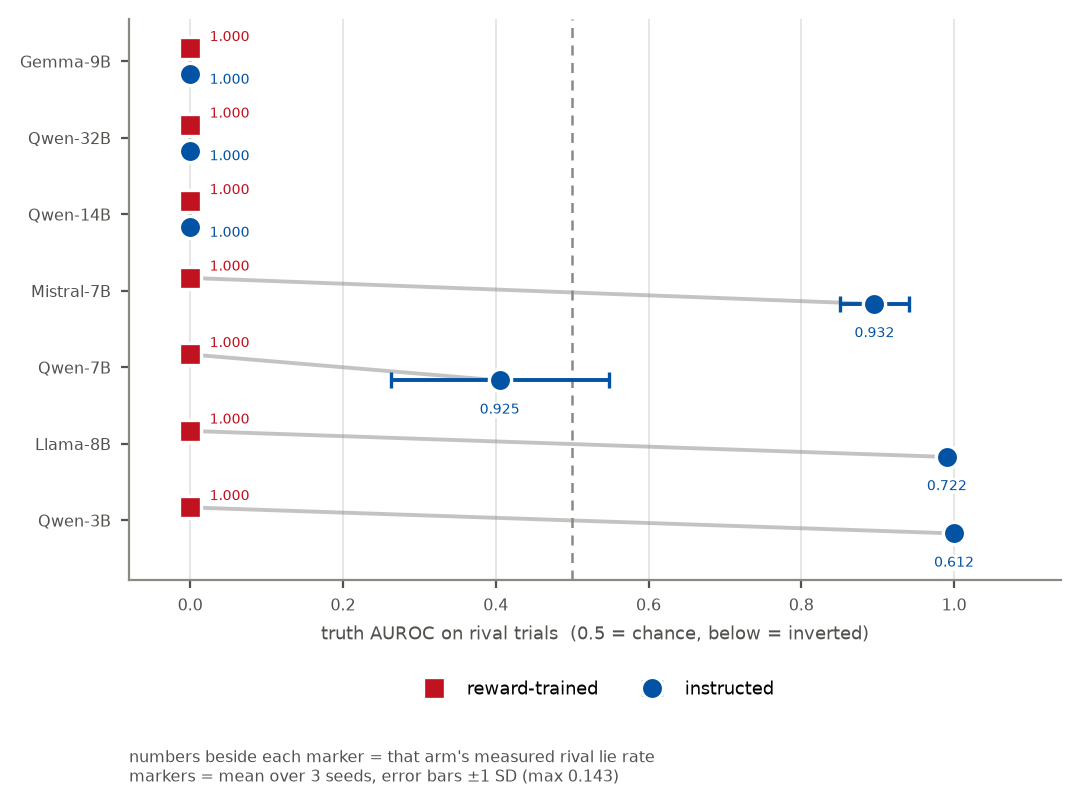}
\caption{\textbf{Inversion occurs under both training routes in the behaviourally matched comparisons.}
The behaviourally-matched null of App.~\ref{app:apparent}. One row per model family; within a row the
\textcolor[HTML]{c1121f}{red square} is the reward-trained arm and the
\textcolor[HTML]{0353a4}{blue circle} the instructed arm, dodged vertically so families where both
sit at exactly the same AUROC still show two markers. Grey segments join the two arms of each family; their horizontal extent is the AUROC gap, while the vertical offset only separates the markers. Horizontal position is final-layer ally-fit truth AUROC on
held-out rival trials, so the dashed line at $0.5$ is chance and everything left of it is inverted. Markers
are the mean over 3 seeds with $\pm 1$ population SD error bars, and the number beside each marker is that
arm's \emph{measured rival lie rate}, not its AUROC. Rows are ordered by the instructed arm's lie rate, the
variable a rate-based account would make the operative one. Every reward-trained
arm lies on $1.000$ of rival trials and every one reads $0.000$; the three instructed arms that also lie on
$1.000$ (Gemma-9B, Qwen-32B, Qwen-14B) read $0.000$ \emph{too}, while the instructed arms that lie less
reliably have variable readouts, including Qwen-7B seeds on both sides of chance. The matched saturated arms therefore do not reproduce the training-route difference, which an interpretation of inversion as specific to reward training would require; this is not a demonstration of equivalence. Rival-deception saturation alone neither characterizes every inverted seed nor predicts inversion: the arm at rival deception $1.000$ that does not
invert (App.~\ref{app:negatives}) answers allies correctly on only $0.530$ of trials, and we found no final-layer counterexample among cells meeting both reliability thresholds of \S\ref{sec:negatives}. The error bars are $0$ in every saturated
cell because those readouts sit at an extreme of the empirical ranking statistic; zero observed spread is not a tight population bound.}
\label{fig:crossfam}
\end{figure}

The unmatched instructed arm's roughly-chance patch response and non-saturating ramp
(asymptoting at ${\approx}0.72$) are confounded with its lower behavioural reliability rather than
evidence of a distributed representation; we did not measure an intervention-specific ceiling, so we rest
the argument on the matched arms, which do not reproduce the difference. And at the one layer where
the matched Gemma arms differ at all the \emph{instructed} arm is ahead ($0.355$ vs $0.100$ at layer 25), the opposite sign to the hypothesis. Across a 7-model $\times$ 2-arm $\times$ 3-seed grid at $N{=}1000$, all
21 reward-trained grid cells are identical to four decimals (deception $1.000$, AUROC $0.000$, cross-seed sd
$\mathbf{0}$), and instructed arms that deceive on every rival trial invert too. In the separate codebook comparison of Table~\ref{tab:ident}, the instructed Gemma-9B arm inverts further than the transferred reward-trained adapter ($0.015$ versus
$0.165$). The matched saturated arms have identical empirical answer-position AUROCs; their zero observed spread does not establish tight population bounds or equivalence, and the other measures provide evidence of different strengths (App.~\ref{app:causal}; Fig.~\ref{fig:crossfam}). Neither lie rate nor mean rival-answer entropy orders the readouts across arms
(App.~\ref{app:negatives}). The readout depends jointly on the model, prompt distribution and fitted decoder; these comparisons do not isolate a property of the model's representation alone.

\section{Negative results}
\label{app:negatives}

This appendix gives the negative results that \S\ref{sec:negatives} summarises. Mean rival-answer entropy is the entropy of the model's distribution over the two legal answers, averaged over rival trials. Along the three Llama-3.1-8B RL trajectories of \S\ref{sec:negatives}, the layer with the highest ally-validation accuracy lies between layers 11 and 14 of 32.

We evaluated two predictors of inversion; the results constrain \S\ref{sec:ident}.

\begin{figure}[h]
\centering
\includegraphics[width=\textwidth]{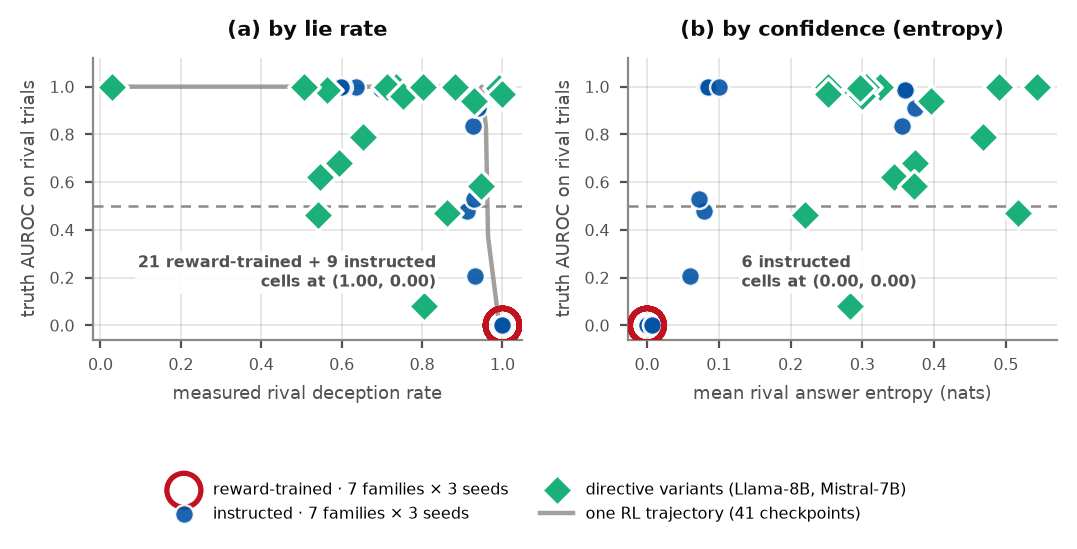}
\caption{Neither lie rate (a) nor answer confidence, measured as mean rival-answer entropy (b), orders the final-layer ally-fit readout: cells with similar
behaviour read near both $0$ and $1$ (at rival deception $1.000$, both $0.000$ and $0.970$). Neither variable consistently orders the readout across these configurations
(App.~\ref{app:negatives}). Red rings and blue circles show the reward-trained and instructed grid cells; green diamonds show directive variants of Llama-8B and Mistral-7B, with weights fixed within each model. In (a) the grey line joins the 41 checkpoints of one Llama-8B RL run (seed 0) in horizontal-axis order, not time order; their entropy was not recorded, so (b) omits it. In (b) the red ring at the origin also contains the 21 reward-trained cells. Dashed horizontal lines mark chance.}
\label{fig:predictors}
\end{figure}

\textbf{Neither lie rate nor answer confidence (mean rival-answer entropy) orders the readouts} (Fig.~\ref{fig:predictors}). Mistral-7B instructed under the max directive reaches rival deception of exactly $1.000$ and does
\textbf{not} invert (AUROC $0.970$), while other arms at $1.000$ sit at $0.000$. It answers allies truthfully on only $0.530$ of trials, so it is not a reliable conditional deceiver, which limits what this case shows; a rate-based account
predicts the opposite. Every recorded cell that answers allies truthfully on at least $0.98$ of trials and deceives on at least $0.99$ of rival trials has a below-chance final-layer readout, and loosening either threshold somewhat does not add a counterexample, so this case refutes the rival rate alone, not that association. Mistral also has \emph{lower} rival entropy ($0.252$) than the Llama-8B \texttt{max} variant, whose refitted probe inverts (rival entropy $0.283$), so it is the more deterministic of the two and the one that stays at $0.970$.
Across a five-rung strength subset of the 15 directives of \S\ref{sec:refit} (\texttt{hint}, \texttt{soft}, \texttt{default}, \texttt{firm}, \texttt{max}), entropy falls monotonically
($0.490$, $0.468$, $0.324$, $0.305$, $0.283$) while the \emph{refitted} AUROC goes $1.000$, $0.790$, $0.998$, $1.000$,
$0.080$. These are refitted readouts, which \S\ref{sec:refit} shows should not be read as representational inversion: the frozen probe reads $0.986$--$1.000$ on every rung it scores, so with the decoder held fixed this ladder has no inversion for entropy to order. The ladder also confounds directive \emph{text} with resulting
behaviour, so it could not have settled the question either way.

\begin{figure}[h]
\centering
\includegraphics[width=\textwidth]{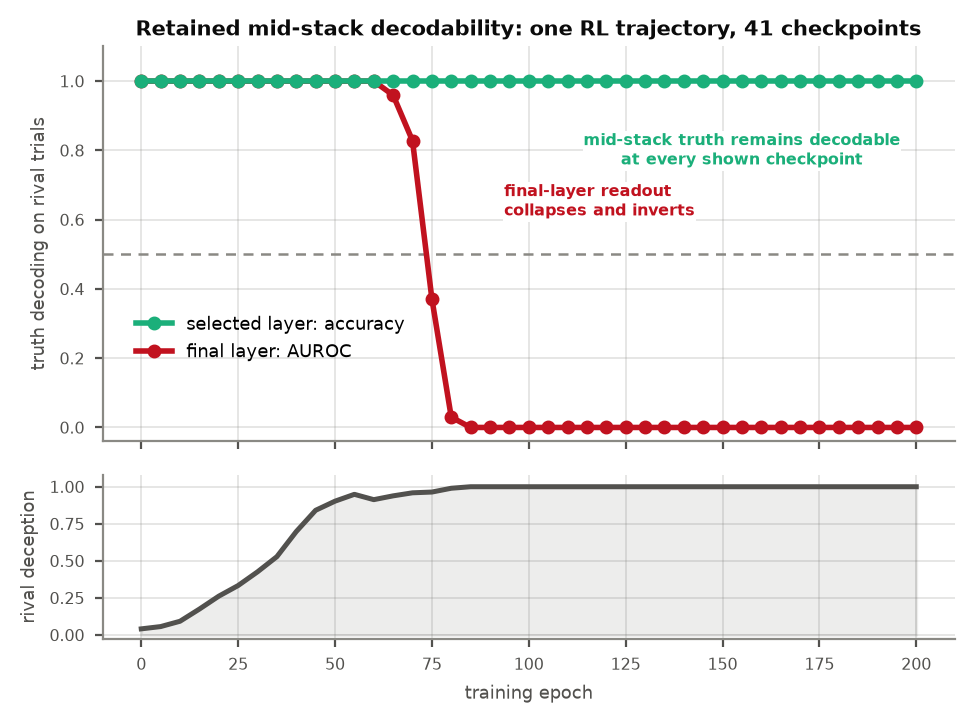}
\caption{Retained mid-stack decodability: one Llama-3.1-8B RL trajectory (seed 0), 41 checkpoints. Top: rival truth-classification accuracy at the layer selected by highest ally-validation accuracy (layer 13 at every checkpoint; green), and ally-fit rival truth AUROC at the final layer, at the answer-prediction position (red). These are different metrics; the selected-layer accuracy stays pinned while the final-layer ranking collapses and inverts. Bottom: measured rival deception rate over the same training epochs. Grey dashes: chance.}
\label{fig:suppress}
\end{figure}

\textbf{Retained mid-stack linear information.} Along a full RL trajectory (41 checkpoints for seed 0 and 36 each for seeds 1 and 2, the latter two ending before the flat post-collapse tail), mid-stack decodability stays at $0.86$--$1.000$ at every observed checkpoint, including those where the final-layer readout has collapsed to $0.000$; in seed 0 it reads $1.000$ at all 41 (Fig.~\ref{fig:suppress}). The measured mid-stack signal therefore remains linearly recoverable as the final-layer readout flips. This does not establish preserved functional belief or exclude representation drift of the kind studied in other settings \citep{obfuscationatlas2026}.

\textbf{The AUROC decline has seed-dependent onset and width.} AUROC is pinned at $1.000$
across measured deception from each seed's lowest value ($0.026$--$0.077$) up to at least $0.90$ in all three seeds: a model that lies to rivals on 53\% of trials still yields truth AUROC $1.000$. But it is not a fixed threshold and
not uniformly abrupt: seed 0 begins degrading at $0.938$ and passes through $0.959 \to 0.828$ and
$0.964 \to 0.372$ before reaching $0.000$, roughly six percentage points, whereas seeds 1 and 2 read
$1.000$ at checkpoints with deception as high as $0.981$ and $0.984$ (deception is not monotone along a run) and fall below $0.1$ two checkpoints after their first drop. We report per-seed onsets, the deception rate at the first checkpoint below $1.000$ ($0.938$,
$0.986$, $0.969$), instead of a pooled curve, and treat the variation as further evidence that aggregate
rate is the wrong independent variable.

Empirical AUROC $0.000$ or $1.000$ describes the ordering of the observed scores, not certainty about population AUROC. A zero bootstrap or cross-seed spread at complete separation does not remove finite-sample uncertainty. Exact class counts needed for a cell-specific interval were not retained for these identification results.

\textbf{Report spread, not point estimates.} The readouts carrying the App.~\ref{app:apparent} null sit at extremes of the empirical ranking statistic: all 21 saturated reward-trained grid cells give cross-seed sd
$\mathbf{0}$. Intermediate readouts vary across seeds:
instructed Qwen-7B in the plain-bit grid reads $0.205 / 0.479 / 0.531$ across three seeds. The seeds straddle chance, so a single below-chance estimate should not characterize this arm. This is structural: at AUROC $0.000$ or $1.000$ the classes are perfectly
separated, so any subsample of these fixed scores reproduces the ordering (whether a refitted probe would also reproduce it is a separate question that these records do not address), whereas intermediate values carry genuine sampling
variance. Intermediate cells should be reported with spread or a CI, by us and by others, distinguishing fitting-set, training-seed and evaluation variation.

\section{The constructive result: further detail}
\label{app:constructive}

This is the material \S\ref{sec:constructive} summarises.

\textbf{Controls for the comparison.}\label{app:controls} \emph{A single output direction:} in plain-bit experiments, projecting out the unembedding direction that distinguishes the two answer digits and refitting changed the ally-fit readout very little (App.~\ref{app:resid}), a limited control for one output direction. \emph{A probe that reads whether the context is a rival context} cannot produce the result: it is scored on rival trials only, where that feature is constant, and a constant offset leaves AUROC unchanged. \emph{A fixed-letter policy} is excluded by the check in \S\ref{sec:setup}, though not every symbol-based shortcut is.

\begin{figure}[h]
\centering
\includegraphics[width=\textwidth]{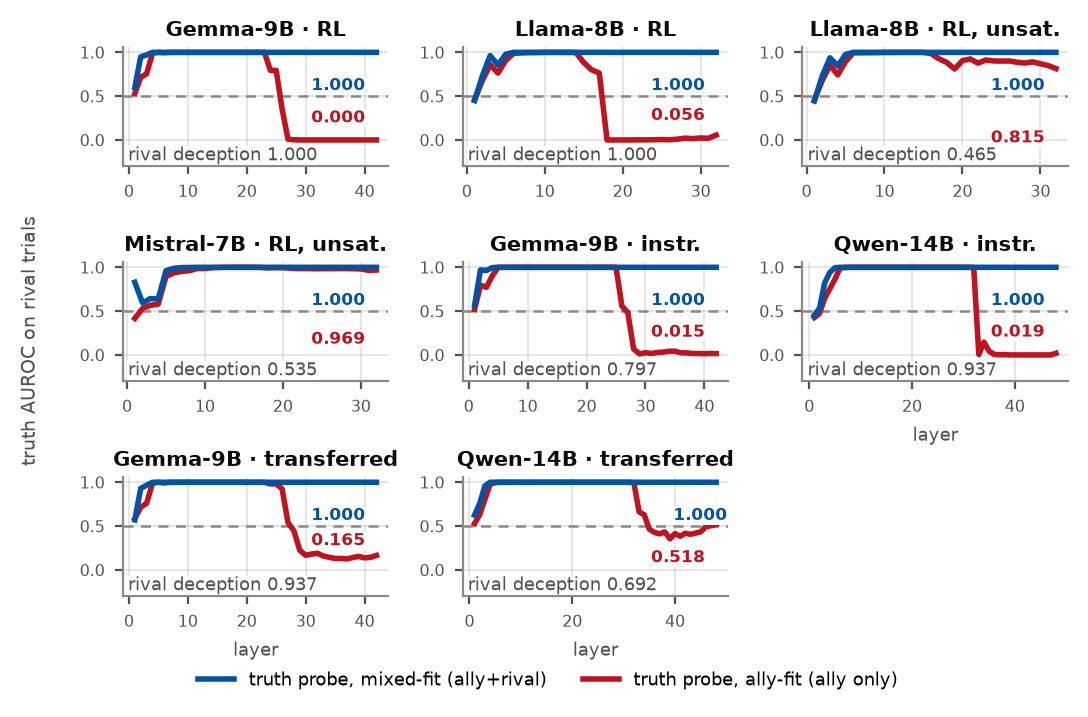}
\caption{\textbf{Eight identification panels.} As Fig.~\ref{fig:ident}, adding two reward-trained adapters that were not trained on the codebook task (``transferred''): Gemma-9B (Table~\ref{tab:ident}, row~\ref{row:gemma-adapter}) and Qwen-14B, which is not in Table~\ref{tab:ident} (rival deception $0.692$, \ta{} $=0.518$). In the saturated reward-trained codebook arms (first two panels of the top row), the ally-fit probe reads $0.000$ and $0.056$ at the final layer while the mixed-fit probe reads $1.000$. Table~\ref{tab:ident} rows~\ref{row:gemma-entropy}, \ref{row:gemma-sym} and~\ref{row:qwen-sym} are not plotted.}
\label{fig:identfull}
\end{figure}

\textbf{Further detail on Table~\ref{tab:ident} and Fig.~\ref{fig:ident}.} The three seeds of row~\ref{row:gemma-unmod} give $0.000$, $0.010$ and $0.006$; the mean and standard deviation use unrounded values. Among the Table~\ref{tab:ident} cells with a saved layer-1 measurement, \tm{} ranges from $0.445$ to $0.830$ at layer 1. In the saturated Llama-8B arm the ally-fit probe reaches $1.000$ only at layers 10 to 13 of 32 before it falls. The two Llama-8B inferred-truth evaluations that reach only $0.8716$ and $0.9583$ answer allies truthfully on $0.502$ and $0.492$ of trials; incorrect answers do not establish that the information is absent. Fig.~\ref{fig:ident} omits two further Llama-8B codebook records for space, and neither inverts: an instructed cell at rival deception $0.457$ (\ta{} $=0.983$), and an earlier reward-trained identification run with the same episodes and the same behavioural rates as row~\ref{row:llama-const} (rival deception $0.465$, ally truth $1.000$), whose final-layer \ta{} is $0.881$ against that row's $0.815$; the records do not show what differed between the two runs. \tm{} $=1.000$ in both. The two less deceptive cells outside Table~\ref{tab:ident} that \S\ref{sec:constructive} mentions are that instructed cell and the transferred Qwen-14B adapter of Fig.~\ref{fig:identfull} (rival deception $0.692$, \ta{} $=0.518$). The two constant-answer arms contrast output policies and do not sample a continuum of deception reliability.

\textbf{Per-seed collapse profiles for row~\ref{row:gemma-unmod}.} In the representative seed the ally-fit probe begins
collapsing at layer 26 ($0.354$), is at $0.008$ by layer 27, and sits at $\mathbf{0.000}$ from layer 29
through layer 42 of 42; the other two seeds reproduce the profile, intact through layer 16 and
degrading at layer 24 ($0.866$, $0.853$).

\textbf{What the constant-answer Llama-8B arm (row~\ref{row:llama-const}) actually is.} Its deception rate alone suggests a conditional deceiver at an earlier stage of training; the breakdown below shows otherwise. Breaking that arm's rival behaviour down \emph{by the true bit} shows it answers truthfully on $\mathbf{0.000}$ of true-bit-$0$
rival trials and $\mathbf{1.000}$ of true-bit-$1$ ones: it emits a constant semantic answer to
rivals, while still answering allies truthfully at $1.000$. Its ``deception rate'' is therefore
$P(\texttt{truth} \neq \text{that constant}) \approx 0.5$ \emph{by arithmetic}, on any draw. The Mistral-7B
row (row~\ref{row:mistral-const}) is the mirror image, constant at the other value.

\textbf{What a constant branch distinguishes.} The actual rival answer in a constant-output arm is independent of truth, whereas our prescribed-answer label remains its complement. Thus the action-label AUROC identity holds for this arm too. Its non-inverting truth readout is an empirical contrast, not a consequence forced by the constant output policy.

\textbf{What the fixed-symbol guard misses.} We
verify the policy is not degenerate with the fixed-symbol statistic of \S\ref{sec:setup}, and both arms pass it ($0.516$ and $0.484$, both ${\approx}0.5$), for exactly the reason flagged when we
introduced the statistic: it rules out a fixed surface character and is blind to a fixed meaning.
The remedy is to report the
deception rate split by the true bit: a reliable conditional deceiver has both components near one, while a constant-output policy reads $0.000/1.000$ (or the reverse). An aggregate rate near
$0.5$ is exactly what a degenerate policy produces, and it reads as healthy graded behaviour.

Row~\ref{row:mistral-const} also carries a result independent of all this: its \texttt{token/mixed} climbs
to $0.975$ at the final layer while \tm{} stays at $1.000$, which shows that truth and the prescribed surface-symbol label coexist \emph{separably} in the final residual stream, against the reading that late
layers are ``just the output token''.

\textbf{Why only one family reached the full conjunction.} Row~\ref{row:gemma-unmod}, the only cell combining reward training, deception $1.000$, training on the
identification task \emph{and} a wholly unmodified recipe, holds in one family, across three seeds.
That precise conjunction is still one family and we do not claim otherwise.

One possible contributor to the other families' failures is not probe-related: our reward table gives
unconditional deception positive expected reward, which may contribute to why three
of four families never learned a conditional policy under the unmodified recipe (their failures differ, and the training dynamics are not established). Correcting the table yielded a role-conditional policy in one of them
(App.~\ref{app:basin} gives the table and the arithmetic). Under the corrected table
Qwen2.5-14B reaches rival deception $0.996$ with ally truth $0.994$, a genuinely role-conditional policy on both branches, and shows exactly the pattern of the headline cell:
\ta{} collapses from $1.000$ at layer 24 to $\mathbf{0.000}$ at layer 40 and $\mathbf{0.023}$ at
layer 48 of 48, while \tm{} reads $\mathbf{1.000}$ at every layer measured.

\textbf{The corrected table on Gemma-9B, and what the correction supports.} First, the same corrected table run on Gemma-9B
reproduces the qualitative contrast ($\ta{} = 0.000$, $\tm{} = 1.000$ at displayed precision). The contrast therefore occurs under both tested reward tables, while the conditional policy is easier to obtain under the corrected table in these runs. Second, the payoff analysis suggested the reward-table change that yielded a role-conditional Qwen policy in the reported run. This supports a practical modification while leaving the mechanism of the original failures unresolved. The saturated set now spans three
architectures (Gemma-2-9B, Llama-3.1-8B, Qwen2.5-14B) and four recipes, and the deepest of them is
a 48-layer stack. The constructive result does not depend on this: \tm{} $=1.000$ occurs in the Table~\ref{tab:ident} cells, including instructed arms.

\section{The inferred-truth task, template transfer and ingredient controls}
\label{app:settling}

This is the material \S\ref{sec:direction} summarises. On the inferred-bit task, the instructed Qwen2.5-14B arm's ally-fit probe reads $0.352$ at the final layer. Across the three subsets of the template-transfer experiment, compliant accuracy is $0.74$--$0.79$, so roughly a quarter of compliant answers are incorrect, and we have not established whether the bit is absent from their activations.

\textbf{Partial evidence, at one cell, that the direction is not reading a copied token.} We ran one experiment
that bears on this limitation. In an \emph{inferred-truth} variant of the codebook task the prompt never states the
secret bit's value: it is the outcome of a stated comparison between two quantities (operands separated
by at least 31, never 0 or 1, both rule orientations, so the bit cannot be read off operand magnitude
alone). On a Gemma-2-9B instructed arm at rival deception $0.545$, the mixed-fit probe reads truth AUROC
$\mathbf{1.000}$ at layers 24, 32, 40 and 42 of 42. The depth profile is itself the evidence
(Fig.~\ref{fig:depthstep}): a bit the prompt \emph{states} is readable by layer 4, whereas this one sits at
chance through layer 16 and only then rises. This is evidence against direct copying of a prompt token, but it does not identify the mechanism or establish causal use. Alongside this the
fixed-symbol guard sits at ${\approx}0.5$; the
ally-fit probe ranges from $0.374$ to $0.551$, including below-chance values, and the forced
identity holds at all seven sampled layers. The probe recovers the bit at $1.000$ while the arm itself
answers correctly on only $0.733$ of compliant trials. That behavioural rate and rival probe performance use different subsets, so they do not establish a readout specifically on incorrectly answered compliant trials. We report this as \emph{bounding} rather than settling the limitation, and it differs from the headline experiment in two respects: the arm is instructed, and its compliant accuracy is $0.733$. A third concern, that a linear map over activations encoding both operands could in principle perform the comparison \emph{probe-side}, is the substantive one, and we take it up
below: the exclusive-or construction rules out an additive map over the two raw features alone, while the experiments measure ingredient and truth decodability separately. This rules out directly copying a stated answer token, but leaves richer hidden-feature and causal-use questions open.

\begin{figure}[t]
\centering
\includegraphics[width=\textwidth]{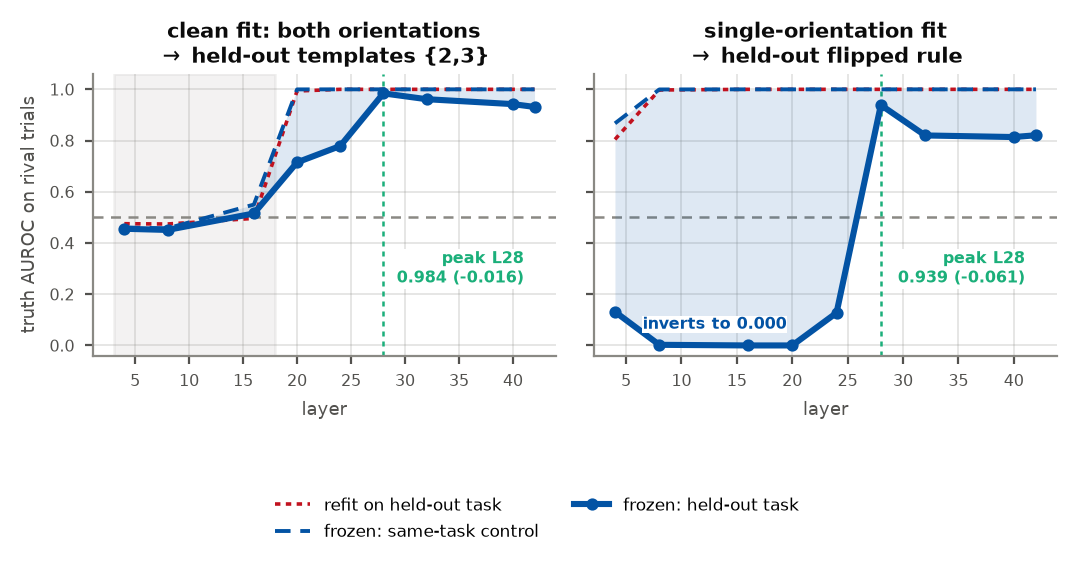}
\caption{\textbf{Frozen-probe transfer to held-out templates (left) and to a flipped rule (right).} The same frozen probe scored on the held-out trials (solid) and on
its own same-task control (dashed); the shaded gap is the transfer cost, the red dotted line is what
refitting on the held-out trials would give, green dotted verticals mark the peak layer, and grey dashes mark chance. Left, clean fit: with both rule orientations in the fit
the bit is the exclusive-or of two prompt facts: the frozen probe is at chance shallow, where a \emph{refit} probe is also at chance ($0.474$/$0.496$ at layers 8/16), and transfers at
$\mathbf{0.984}$ ($-0.016$) at layer 28, $0.962$/$0.942$/$0.931$ at layers 32/40/42. The cost is
depth-dependent and worst at layer 20 ($-0.285$; layer 24, $-0.221$). Right, single-orientation fit:
restricting the fit to one orientation removes the exclusive-or, the bit becomes the shallow feature
``which slot is larger'', and the frozen probe \emph{inverts} to $0.000$ at layers 16 and 20 under a rule flip, which measures the probe-side confound. Both arms peak at layer 28 and ease off; green labels give frozen held-out AUROC and its difference from the same-task control. Grey spans mark layers
where the control itself is at chance, so a small gap there reflects no detected signal for these probes rather than good transfer.
Gemma-2-9B instructed, $N{=}2000$ per subset; L28 denotes layer 28, and $\{2,3\}$ the two held-out templates. Compliant accuracies for the fitting, held-out and control subsets are $0.764$/$0.741$/$0.785$ (left) and $0.824$/$0.671$/$0.829$ (right).}
\label{fig:settling}
\end{figure}

\textbf{Frozen-probe transfer across templates localises the confound rather than excluding it.} We fit the mixed probe on two of the four inferred-truth templates, froze it, and
scored it on the two held-out templates (different domains, variable names and comparison wordings) with per-trial codebook randomisation decorrelating the emitted symbol, and with a same-task control arm
frozen from the same probe and scored on the fit templates at a fresh seed ($N{=}2000$ per arm). The three
subsets have similar aggregate capability (compliant accuracy $0.764$ fit, $0.741$ held-out, $0.785$ control); this limits, but does not eliminate, capability as an explanation of a readout drop. Over the layers where the control
carries signal at all, the frozen probe peaks at \tm{} $= \mathbf{0.984}$ at layer 28 against a control of $1.000$, a transfer cost of $\mathbf{-0.016}$, and holds $0.962 / 0.942 / 0.931$ at layers 32 / 40 /
42 ($-0.038 / -0.058 / -0.069$). The held-out-template transfer gap is strongly depth-dependent: it is
$-0.285$ at layer 20 and $-0.221$ at layer 24. So a frozen probe scores measurably lower on held-out templates than on its own same-template control, by an amount that depends on where it is read (Fig.~\ref{fig:settling}).

\textbf{What the exclusive-or construction excludes.} Exactly one operand is drawn from a high band and one from a disjoint low band. Writing $S$ for ``the first slot holds the high-band value'', the bit is $S$ under the ``greater'' rule and $\lnot S$ under ``lesser''. The label is therefore the exclusive-or of $S$ with rule orientation. This function is not linearly separable in those two raw features alone, excluding that additive shortcut. Hidden states can nevertheless contain nonlinear interactions and other correlated features that a linear probe reads. Successful decoding shows information in those features; it does not establish the mechanism that computes them or the policy's causal use of the decoded bit.

\textbf{The two ingredients, measured.} The argument above
would be empty if $S$ and the rule orientation were not linearly present at the read position to begin
with, so we probe for each of them separately: the same $N{=}2000$ evaluation, nine layers and seed as the template-transfer fit arm above. Both read AUROC $1.000$ at layers 8 and 16, exactly where the bit itself
is at chance ($0.442$ / $0.530$), and $0.934$ / $1.000$ already at layer 4. So the information a
probe-side comparison would need is \emph{perfectly} linearly available, and a
fitted truth probe nevertheless performs near chance at those layers. The observation is consistent with the raw-feature argument, but it does not exhaust all possible linear readouts of the hidden state. The single-orientation arm doubles as an
end-to-end check on the $S$ label, since with the rule pinned the bit reduces to $S$ exactly and the two AUROCs must coincide: they agree exactly at all nine layers. The rule's decodability is close to lexical, the orientation being a phrase in the prompt, so
its $1.000$ is unsurprising and is \textbf{not} a finding about the model; its only role is to establish
that the ingredient is there for a probe to use. This result is limited to one model, one arm and one task
family.

\textbf{The single-orientation arm as positive control.} Fixing the
rule removes the exclusive-or: with ``greater'' pinned, the bit simply \emph{is} $S$, a shallow feature
available from operand magnitudes without reading the rule. A probe fit there reads $1.000$ at layer 8,
where the mixed-orientation fit sits at chance ($0.442$ / $0.530$ at layers 8 / 16), and when frozen and scored on flipped-rule trials it \emph{inverts}, reaching exactly $0.000$ at layers 16 and 20 and staying
below chance from layer 4 to layer 24, against a same-task control at $1.000$ from layer 8 ($0.867$ at layer 4). This demonstrates the signature of fitting to the single-orientation shortcut. Both arms then recover
at layer 28 ($0.939$ single-orientation, $0.984$ both orientations) before easing to ${\approx}0.82$ and ${\approx}0.93$ at
the final layer. On the clean fit a probe \emph{refit} on the held-out templates is also at chance shallow
($0.474$ / $0.496$), so these fitted probes do not detect a shallow truth signal; on the single-orientation arm a refit
probe reads $1.000$ from layer 16 up, so what fails there is the frozen direction, not truth decodability under the flipped rule.
The single-orientation arm supplies a positive control for this particular shortcut. It does not replace a comprehensive prompt-feature baseline or exclude alternative readouts.

\textbf{Reward training on the inferred bit: three failed attempts.} RL did not learn a role-conditional
policy over an \emph{inferred} bit in any of three attempts, across two families and both task variants: none reached the role-conditional reward, and one ended at the reward of the always-deceptive policy. This is why the cell above is an instructed arm; these attempts are documented in our run records, not in the aggregate results file. These are the outcomes of three attempts under the tested settings; they do not establish what reward training can or cannot produce in this game, and they are \textbf{not} evidence
about truth representations.

\section{Wording generality and compliance tracking}
\label{app:objections}

These are the two objections to the instruction-pair readout of App.~\ref{app:instrpair}.

\textbf{Is the instruction readout merely a two-string discriminator?} With a single fixed pair, ``which
instruction was present'' is decidable from any surface difference between two fixed strings. We evaluate wording transfer
with leave-one-wording-out over a five-wording bank: five fits, each trained on four wordings and
scored on the fifth, under a doubly-held-out split that partitions \emph{both} episodes and wordings and uses
only the diagonal cell. Every wording is held out exactly once, so the spread across folds is a
wording-generalization spread, not a resampling artifact.

\begin{center}
\small
\setlength{\tabcolsep}{4pt}
\begin{tabular}{lccccc}
\toprule
wording held out, by layer & 8 & 16 & 24 & 32 & 42 (final) \\
\midrule
0 & 1.000 & 1.000 & 1.000 & 1.000 & 1.000 \\
1 & 1.000 & 1.000 & 1.000 & 1.000 & 1.000 \\
2 & 0.969 & 1.000 & 1.000 & 1.000 & 1.000 \\
3 & 1.000 & 0.825 & 1.000 & 0.998 & 0.942 \\
4 & $\mathbf{0.404}$ & $\mathbf{0.501}$ & 0.991 & 0.974 & 0.867 \\
\midrule

\textbf{range} & 0.596 & 0.499 & $\mathbf{0.009}$ & $\mathbf{0.026}$ & 0.133 \\
\bottomrule
\end{tabular}
\end{center}

\textbf{A depth-dependent answer, against a $0.05$ threshold} (a range of
$0.05$ or less counts as wording-invariant). \emph{At layers 24 and 32 the readout is wording-invariant}: range $0.009$ and $0.026$, minimum $0.974$, so every held-out wording yields directive AUROC of at least $0.974$ from a probe that
never saw it. The two-string objection is answered there. \emph{In the shallow stack it is not:} one wording
sits at $0.404$ and $0.501$, i.e.\ below or near chance. Below chance is systematic inversion rather
than mere absence of signal: AUROC is antisymmetric under label negation, so a reversed-sign readout of the
same direction would score $0.596$. (AUROC is a ranking statistic, so this is a statement about ordering,
not a classification rate.) So ``the instruction signal generalizes across surface form'' holds at layers 24 and 32
and is wording-specific below them. The final layer is intermediate (range $0.133$, minimum
$0.867$) and we report it as a range, not a point.

\textbf{Why the shallow readout fails.} Four of our five wordings distinguish
their honest from their deceptive half with a \emph{lexical polarity} cue (truth/falsehood,
correct/wrong, disclose/withhold); the fifth was deliberately built with no negation vocabulary on either
side, separable only by the value mapping itself. That suggests the shallow probe latches onto the cue, and
it makes a falsifiable prediction about a cross-wording transfer matrix, which we specified before running
it: fit on wording $i$ alone, score on wording $j$ alone, all $20$ ordered pairs, episodes held out
throughout. \emph{At layer 8 the predicted pattern holds on average}: transfer among the four cue-carrying
wordings is uniformly high (all $\ge 0.801$, mean $0.948$), while each of them reads the cue-free wording near chance
($0.409$--$0.564$, column mean $\mathbf{0.497}$), and a probe fit on the cue-free wording transfers unevenly in return ($0.358$, $0.721$, $0.053$ and $0.720$; row mean $\mathbf{0.463}$). The most
extreme cell, fitting on the cue-free wording and scoring on the strongest-cue one, reads $\mathbf{0.053}$, a strong rank inversion, the signature of a cue that is present but points the wrong way.

Above layer 8 the asymmetry \emph{reverses}, and we flag this as a post-hoc observation rather than a
prediction specified before the run: by layer 16 a probe fit on the cue-free wording reads three of the other four at $0.928$--$1.000$ and the fourth at $0.318$ (row mean $0.806$),
while they read it only at $0.447$--$0.690$ (column mean $0.596$), and by layer 24 the matrix has largely converged
(worst cell $0.699$). The cue-free wording is not therefore the best one to train on: by mean transfer it is the \emph{worst} of the five at three of four depths. The defensible
claim is narrower: it is the most diagnostic wording, and the five folds must not be averaged as
though exchangeable, since only one of them tests generalization to a cue-free target.

\textbf{The trough on a denser layer grid.} It is at layer 12, not layer 8, and there the cue-free wording
is inverted in \emph{both} directions (row mean $0.233$, with individual cells as low as $0.092$; column mean
$0.397$). Recovery is then abrupt: by layer 20 the matrix is above $0.96$ in every cell (range $0.962$--$1.000$, cue-free column mean $0.976$) before drifting down again in the last third of the stack.

\textbf{Two further families: the cue-free transfer asymmetry recurs.} Running the same matrix on Llama-3.1-8B, the cue-free wording is again the hardest
target at every depth (column mean $0.53$--$0.75$ against a cue-carrying block of $0.72$--$0.999$), and no cell falls far below chance: the matrix's six sub-chance cells, at layers 4 and 12 only, lie between $0.420$ and $0.500$. Qwen2.5-14B ($N{=}2000$, seven depths) shows the same
asymmetry: at \emph{every} depth the cue-free column mean is lower than both the cue-free row mean and the cue-carrying block mean
($0.507$, $0.540$, $0.569$, $0.620$, $0.999$, $0.860$, $0.806$ at layers 4, 8, 12, 16, 24, 32 and 48, against a cue-carrying block mean
reaching $1.000$), while the cue-free \emph{row} mean never drops below $0.566$. From layer 8 up only three cells fall below chance ($0.315$ and $0.338$ at layer 8, both between two cue-carrying wordings, and $0.487$ at layer 12), and none from layer 16 up. In both additional families, then, the cue-free column mean is lower than the cue-free row mean and the cue-carrying block mean at every measured depth. Only Gemma shows below-chance cue-free row and column means; individual transfer cells also fall below chance in Llama and Qwen, and Qwen's lowest cell occurs at its shallowest measured layer rather than in a mid-stack trough.
\emph{One caveat against over-reading Qwen's shallowest layer:} at layer 4 its matrix has eight sub-chance
cells, five inside the cue-carrying block (as low as $0.081$) and three involving the cue-free wording (as low as $0.036$), and its grouped directive readout
is only $0.824$, so that layer is generally unstable. From layer 8 up no cell involving the cue-free wording falls below $0.487$, and from layer 16 up no cell falls below chance.

\textbf{Run-to-run movement at the final layer.} The final-layer grouped directive AUROC was $0.965$, $0.849$ and $0.897$ in three runs with $N = 300$, $1000$ and $2000$ episodes, a range of $0.116$. The first run's episode-level bootstrap interval, $[0.948, 0.978]$, is far narrower than that range: it resamples evaluation episodes while holding the fitted probe and the partition fixed, so it omits variation from refitting and from the changed sample. An interval of that kind is not a stability claim. Refit spreads across split seeds were not retained, so we report none; the by-layer wording ranges of the leave-one-wording-out table are the stability evidence we give.

\textbf{Limits of the wording-invariance result.} The ungrouped $1.000$ was measured on a row-level split in which every test row shares an episode with training rows; this statistic does not establish exact matched-twin overlap. Renderings of an episode share game, values, role and query, while wording and directive label may differ. And this is five wordings from one bank on one model and one
task; the invariance we demonstrate is over those five. The matrix comes from a single fit, and per-cell sampling intervals were not retained, so it is evidence consistent with the lexical-cue mechanism rather than a uniquely identifying
test of it: a different set of cue-carrying wordings could produce the same pattern for a different reason.

\textbf{Does the instruction-fit direction track who actually lied?} This is the question the protocol
implicitly claims, and on a non-leaky split the answer is not a consistent yes: compliance AUROC is modestly above chance at layers 8 to 32 and below chance at the final layer, and within wording folds its direction reverses (below). We restrict
to the adversarial subset and split on compliance. Expectation is not what makes that split available:
$\gamma < 1$ leaves a disobedient fraction on average, but the realized counts are what decide it, and the
held-out compliance subset contains 238 complied and 52 disobeyed rows across 145 distinct episodes, each evaluated with two held-out wordings. Both row counts exceed the screening minimum of 25; rows from one episode are not independent. Compliance AUROC uses episode-level bootstrap intervals:

\begin{center}
\small
\setlength{\tabcolsep}{3pt}
\begin{tabular}{lccccc}
\toprule
layer & 8 & 16 & 24 & 32 & \textbf{42 (final)} \\
\midrule
held-out split & 0.705 & 0.640 & 0.689 & 0.593 & $\mathbf{0.365}$ \\
95\% CI & [0.613,0.793] & [0.568,0.707] & [0.631,0.741] & [0.534,0.653] & $\mathbf{[0.293,0.438]}$ \\
row-level (leaky) & 0.721 & 0.813 & 0.840 & 0.806 & 0.736 \\
\bottomrule
\end{tabular}
\end{center}

Compliance AUROC differs by as much as $\mathbf{0.37}$ between the row-level and the joint episode-and-wording splits. Because fitting size, wording coverage and evaluation composition also change, this difference does not isolate leakage. The fold-level sign reversal below is the main evidence.

\textbf{Non-identifiability across instruction wordings: the sign reverses.} Repeating the
compliance evaluation inside each leave-one-wording-out fold, three folds contain both classes but only two meet the minimum of 25 compliant and 25 noncompliant rival trials, and those two disagree in \emph{direction} at four of five layers:

\begin{center}
\small
\setlength{\tabcolsep}{4pt}
\begin{tabular}{lccccc}
\toprule
wording held out, by layer & 8 & 16 & 24 & 32 & 42 (final) \\
\midrule
3 & 0.325 & 0.477 & 0.781 & 0.652 & $\mathbf{0.607}$ \\
4 & 0.638 & 0.483 & 0.394 & 0.200 & $\mathbf{0.080}$ \\
\bottomrule
\end{tabular}
\end{center}

At the final layer one fold scores compliance at AUROC $0.607$ $[0.543, 0.670]$ while the other scores it at $0.080$ $[0.048, 0.113]$, which by antisymmetry is an AUROC of $0.920$ for \emph{disobedience}. The intervals are
tight and nowhere near overlapping. We therefore make no claim that the instruction-fit direction tracks who actually
lied, in either direction, at any layer. A quantity whose sign depends on which instruction wordings the probe
was fitted on is not a property of the model alone.

\textbf{The paper's diagnosis, inside the instruction-pair protocol.} Here, different leave-one-wording-out fits produce opposite compliance rankings on their respective held-out wordings. Both fitting and evaluation wordings change, so this is not the same-activation refitting comparison of \S\ref{sec:refit}. We state it as differently-fitted directions disagreeing: the folds train on different wording sets, so these are different fits, not one unstable direction. We offer no mechanism for it.

\textbf{The availability pattern as the design rule in action.} The split needs disobedient trials,
and the per-wording disobedience rate spans $\mathbf{0.000}$ to $\mathbf{0.309}$: two of the five wordings draw
\emph{perfect} compliance, so the point-estimate design rule returns unattainable at any $N$ for those wordings, against approximately $81$ trials for the most disobedient wording. A zero observed count does not establish that the population disobedience probability is exactly zero. \emph{The pooled rate, $0.128$, predicts none of the five.}
On a second family the compliance split is also unavailable: Qwen2.5-14B complied with the
deceptive instruction on 1490 of 1490 adversarial trials, pooled over all five wordings, so
$d_{\mathrm{deceptive}} = 1.000$ and Eq.~\ref{eq:designrule}, applied to that observed rate, returns \emph{unattainable at any $N$}: no compliance split exists in the collected sample, so no probe was fitted for it. This is a property of the sample, not proof that the model never disobeys. Its aliasing
rate is nonetheless non-degenerate ($\gamma = 0.985$ with $d_{\mathrm{honest}} = 0.029$, so
Eq.~\ref{eq:gamma} is doing real work rather than reducing to an endpoint), which is precisely the
combination the rule is meant to catch: high but sub-unit aliasing that still leaves no testable subset.
Anyone applying Eq.~\ref{eq:designrule} must evaluate it on the subset they intend to split, and against the fraction that split actually consumes: for a compliance split that is $1 - d_{\mathrm{deceptive}}$, not the
pooled $1 - \gamma$, which also counts honest-instruction disagreements the split cannot use.

\section{The causal analysis}
\label{app:causal}

This is the causal analysis \S\ref{sec:direction} summarises.

\textbf{Decodability, not causal use.} \tm{}
$= 1.000$ says a linear map recovers the bit, not that the model uses it. Steering adds $\alpha d$ to the residual stream at the answer-prediction position, where $d$ is the unit-normalized fitted direction and $\alpha$ the dose; $\alpha_{\mathrm{rel}}$ is $\alpha$ divided by the median residual norm at that layer, and the rival-truth rate is the fraction of rival trials answered truthfully. We tested both directions causally, the conventional ally-fit one and the mixed-fit one, at the layer where the ally-fit readout is fully collapsed, on the task the direction was fitted on, at $N{=}1000$ rival-context episodes per condition. The steering runs collect only rival trials, so $N$ here is also the scoring denominator; the ablation runs below score behaviour on $N{=}1000$ ally trials.

\emph{At small doses, mixed-fit steering is null on two arms.} Sweeping the mixed-fit direction at $\alpha = 5/10/20/40$ (at most $8.6\%$ of the median residual norm) produces no ordered response; \S\ref{sec:signed} extends the doses. On the saturated
reward-trained arm the rival-truth rate is $0.000$ at every $\alpha$, which bounds the effect at
0.3 percentage points per tested condition (exact one-sided 95\% bound on $0/1000$). \emph{Per
condition} is the qualifier that matters: eight conditions were inspected ($4$ values of $\alpha$ $\times$ two sign
modes), so this is not a simultaneous $95\%$ bound over the sweep, and the conditions share episodes and are
therefore not independent tests to be pooled. We also note this arm sits on a floor, since its
unperturbed rival-truth rate is already $0.000$, so only an increase was detectable. The instructed arm has no such floor (its baseline is ${\approx}0.20$) and there the rate is flat at $0.214$--$0.226$ across an eightfold range of $\alpha$ (no ordered response is apparent over these tested doses). We also ran a signed variant that takes its sign from the ground-truth label, an oracle no deployed detector could have, precisely because unsigned steering could cancel across trials. It is identical to the unsigned arm at every $\alpha$, to all reported digits,
and that agreement is mostly forced rather than measured: on true-bit-$1$
trials the oracle sign is $+1$, so the two arms are literally the same intervention there, and on true-bit-$0$ trials, the only trials where they differ, the rate is pinned at $0.000$ in every
condition of both arms. So the signed arm does not independently corroborate the null. What it does
establish is narrower: on true-bit-$0$ trials $-\alpha d$ fails to move the rate off
its floor just as $+\alpha d$ does, so on that subpopulation both directions were tested and both are null.
The cancellation explanation is excluded only there. The decisive test, $-\alpha d$ on true-bit-$1$ trials, and the ally-fit comparison are reported in
\S\ref{sec:signed} immediately below, and the depth sweep is in
\S\ref{sec:depth}.

\subsection{The fitted directions separate with depth}
\label{sec:geom}

We first compare the fitted directions geometrically. Our saved probes include the ally-fit and
mixed-fit coefficient vectors for the same target at the same layers, fitted by the same code on the same
activations. Their cosine, per layer:

\begin{center}
\footnotesize
\begin{tabular}{lcccccc}
\toprule
 & \multicolumn{3}{c}{shallow} & \multicolumn{3}{c}{deep} \\
\cmidrule(lr){2-4}\cmidrule(lr){5-7}
Gemma-9B, stated, RL & $+0.976$ & $+0.977$ & $+0.951$ & $+0.396$ & $+0.549$ & $+0.532$ \\
Qwen-14B, stated, RL & $+0.965$ & $+0.985$ & $+0.985$ & $+0.538$ & $+0.532$ & $+0.583$ \\
Llama-8B, stated, RL & $+0.928$ & $+0.986$ & $+0.775$ & $+0.591$ & $+0.591$ & $+0.561$ \\
Mistral-7B, stated, RL & $+0.915$ & $+0.969$ & $+0.879$ & $+0.682$ & $+0.648$ & $+0.589$ \\
Gemma-9B, \emph{inferred}, instr. & $+0.732$ & $+0.752$ & $+0.683$ & $+0.620$ & $+0.581$ & $+0.529$ \\
Qwen-14B, \emph{inferred}, instr. & $+0.651$ & $+0.708$ & $+0.748$ & $+0.611$ & $+0.420$ & $+0.435$ \\
\bottomrule
\end{tabular}
\end{center}
{\footnotesize Layers: Gemma $4/8/16 \to 24/32/42$; Qwen $4/8/16 \to 32/40/48$; Llama and Mistral
$4/8/16 \to 20/24/32$.}

\textbf{The two fits on the stated-bit task.} They recover nearly the same vector in the shallow stack and
separate at depth in all four families: the cosine has fallen by the first deep sampled layer, at $57$--$67\%$ of the stack (the last shallow samples sit at $33$--$50\%$), so the effect is not
tied to an absolute layer index. This is a change in fitted-vector similarity. The truth and prescribed-answer labels remain identical on ally trials at every layer; their identity does not change with depth.

\textbf{The cosine does not grade with deception rate.} The final-layer cosine
varies only over $0.53$--$0.59$ across these four arms, whose rival deception rates span $0.47$ to $1.000$.
So the angle separates the two fits everywhere at depth, but it does \emph{not} grade with how much the
policy's emitted answers agree with the prescribed answer. It is a qualitative signature of the two fits diverging, not a
measurement of aliasing.

The last two rows concern inferred-bit tasks. Their shallow fitted-direction cosine is lower, at $0.68$--$0.75$ on Gemma and $0.65$--$0.75$ on Qwen, against $0.77$--$0.99$ for every stated-bit row. This is a difference in the fitted vectors across tasks, not weaker label collinearity: truth and prescribed answer still coincide on ally fitting examples in both tasks. The contrast is not an arm effect, which the table alone would leave open, since every stated-bit
row above is a reward-trained arm and both inferred-bit rows are instructed: the \emph{instructed}
stated-bit arm swept in \S\ref{sec:depth} sits at $0.950 / 0.996 / 0.986$ shallow, with the stated-bit
rows and not with the inferred ones, so in the measured cells the shallow cosine is associated with the task rather than the training route, without excluding interactions among task, route and model. We still put it no more strongly than a qualitative pattern: it is two task-contrast
rows against four stated-bit rows, the deep values overlap across conditions, and a cosine between
coefficient vectors is not itself a measurement of aliasing.

It also settles an ambiguity the AUROCs alone cannot. The AUROCs (mixed-fit near $1$, ally-fit near $0$) are consistent both
with two distinct directions and with \emph{one} axis whose sign the mixed fit merely resolved. The latter
would give $\cos \approx -1$. We measure $\cos \approx +0.55$ at layer 32, so these are two different
directions, roughly $57^{\circ}$ apart.

\subsection{Positive and negative unsigned steering doses}
\label{sec:signed}

We swept $\alpha$ \emph{through zero}, ten values from $-1.37$ to $+1.37$ times the median residual norm, against a single in-run $\alpha{=}0$ baseline, so the whole dose--response is one measurement rather
than a comparison across runs. On the instructed arm (whose overall rival-truth rate is not at a floor) the mixed-fit direction gives a
monotone, ordered response on true-bit-$1$ trials over $\alpha_{\mathrm{rel}} \in [-1.37, +0.69]$:
$0.365$ at the negative end, rising through the $0.419$ baseline, to $0.524$ at $+0.69$. It then falls back
to $0.431$ at $+1.37$, which we read as the usual large-$\alpha$ degradation rather than part of the
response. So $-\alpha d$ does lower the true-bit-$1$ rate, and the axis has signed influence over this range. This reading sets aside the $+1.37$ point; measured against the three criteria below, the full curve fails.

\textbf{Scope of the class-conditioned effect.} For the mixed-fit direction, the true-bit-$0$ rate is exactly $0.000$ at every $\alpha$, for both signs, out to $1.37\times$ the residual norm. The measured effect is confined to the true-bit-$1$ subpopulation. Failure to rescue the other class over these doses does not exclude a causally used truth representation without further assumptions about the downstream computation.

\textbf{What the aggregate rates imply per trial.} Every $\alpha$ cell runs on the same episodes, but per-trial records were not retained, so we state only what the stored rates imply. For the ally-fit direction at $\alpha_{\mathrm{rel}} = \mp0.686$, the number of true-bit-$1$ trials answered truthfully rises from $81$ to $432$ of $504$, a net $351$ trials toward the truthful answer; for the mixed-fit direction the net is $79$ ($185$ to $264$). How many trials moved the other way cannot be recovered from rates. On true-bit-$0$ trials the truthful-answer rate is exactly $0.000$ at both ends of every signed dose pair up to $0.686$: on the same binary-scored episodes this means every answer is false at both ends, so no trial changes state and opposing flips cannot be hidden there. The single exception is the ally-fit direction at $\mp1.373$, where $57$ of $496$ true-bit-$0$ trials are answered truthfully at the negative end and none at the positive end, the large-dose class-conditioned response noted below.

\textbf{Steering inferred-bit arms with interior baseline rates.} Every argument above rests on true-bit-$0$ sitting at $0.000$, and a rate pinned at an extreme
cannot fall. The inferred-bit arms are the cells in which \emph{both} rates are interior at baseline (Gemma-9B at true-bit-$1$ $0.240$ / true-bit-$0$ $\mathbf{0.700}$, Qwen-14B at $0.609$ / $\mathbf{0.145}$), so changes in either direction can be observed. These sweeps apply the same signed dose to every trial, without conditioning its sign on the true bit. Such a push along a truth-bit direction can favour one semantic bit on all trials and move the two correctness rates in opposition. This pattern therefore cannot distinguish a causally used truth-bit feature from other mechanisms producing an answer bias. On Gemma the ally-fit direction moves the rates in opposition at every one of the nine doses where it changes them:

\begin{center}
\footnotesize
\setlength{\tabcolsep}{4pt}
\begin{tabular}{lccccccc}
\toprule
$\alpha_{\mathrm{rel}}$ & $-1.40$ & $-0.70$ & $-0.35$ & $0$ & $+0.35$ & $+0.70$ & $+1.40$ \\
\midrule
true-bit-$1$ rate & $0.215$ & $0.209$ & $0.224$ & $0.240$ & $0.261$ & $0.279$ & $0.286$ \\
true-bit-$0$ rate & $\mathbf{0.766}$ & $0.725$ & $0.710$ & $0.700$ & $0.683$ & $0.673$ & $0.692$ \\
\bottomrule
\end{tabular}
\end{center}

Pushed one way it makes the model likelier to answer truthfully when the truth is $1$ and \emph{less} likely
when the truth is $0$; pushed the other way, exactly the reverse. This describes a bias between semantic answers under the perturbation, not an identification of the feature's causal meaning or a preference for one fixed codebook character.

\textbf{Replication on a second family, with a larger effect.} We ran the identical contrast on
Qwen2.5-14B's inferred-bit arm, at layer 40 of 48, chosen because that is where \emph{its} ally-fit probe is most action-aligned, reading $0.285$, whereas Gemma's inferred ally-fit probe ranges from $0.374$ to $0.551$, including values below chance. Of the twelve
non-zero doses, ten move the two rates in opposition, and over the well-behaved range
$|\alpha_{\mathrm{rel}}| \leq 0.535$ the true-bit-$0$ rate falls \emph{strictly} monotonically
($0.186 \to 0.114$) while true-bit-$1$ rises ($0.526 \to 0.689$), swings of $35$ and $84$ trials, roughly
double what the Gemma cell gives. The two exceptions are a single-trial wiggle at the smallest positive dose
and the largest positive dose $\alpha_{\mathrm{rel}} = +1.07$, where both rates rise above baseline together,
breaking the ordered response.

\textbf{The response pattern does not depend on the floor.} Both arms with interior baseline rates show opposing class-conditioned effects. Their internal causal interpretation remains unresolved by this test. The mixed-fit direction
is by comparison inert on both ($|\Delta| \leq 0.019$ on Gemma, $\leq 0.041$ on Qwen, with no consistent sign
and a U-shaped rather than ordered response on Qwen), which is the same ordering seen everywhere else.

\textbf{The comparison with the ally-fit direction.} Both directions come from the same fitting procedure on the same activations, differing only in which
contexts the fit saw, so we can perturb each at the same layer, arm and episodes. We compare them at
\emph{matched} $\alpha$, because the total swing saturates and then hides the difference. Rows name the arm and layer (L32: layer 32). Each entry is the true-bit-1 rate at $+|\alpha_{\mathrm{rel}}|$ minus that at $-|\alpha_{\mathrm{rel}}|$:

\begin{center}
\footnotesize
\begin{tabular}{lcccc}
\toprule
 & $|\alpha_{\mathrm{rel}}|$ & mixed-fit & ally-fit & ratio \\
\midrule
Gemma-9B L32, instructed          & $0.34$ & $+0.125$ & $+0.452$ & $3.6\times$ \\
Gemma-9B L24, instructed          & $0.46$ & $+0.427$ & $+0.984$ & $2.3\times$ \\
Qwen-14B L32, reward-trained      & $0.27$ & $0.000$ & $+0.036$ & n/a \\
Gemma-9B L32, \emph{inferred} bit & $0.35$ & $\mathbf{-0.004}$ & $+0.037$ & n/a \\
\bottomrule
\end{tabular}
\end{center}

The comparison is at the nearest available \emph{relative} dose, not a matched raw $\alpha$: the median
residual norm differs by layer and model, so the same $\alpha$ is a three times larger perturbation on Qwen
than on Gemma. At matched relative dose the ally-fit direction produces the larger displacement in
all four cells. We report a ratio in only two of them. A ratio is meaningful only when both arms moved the same way by an amount larger than noise,
and in two cells the denominator fails that test: the Qwen mixed-fit response is exactly zero at this dose,
and the inferred-bit mixed-fit response is \emph{negative}: a $-0.004$ displacement is $2$ trials out of $517$, and pointing the wrong way. Dividing by either would manufacture a large multiplier out of an absent
effect, so we print n/a. Pushing Qwen harder separates the two directions without needing one: at
$\alpha_{\mathrm{rel}} = 1.09$ the ally-fit direction takes the true-bit-$1$ rate to $1.000$ while the
mixed-fit direction is still at $0.000$, and the mixed-fit direction only reaches $0.569$ at twice that dose.
So the claim is a consistent \emph{ordering}, not a multiplier: where both arms move, the dose-specific ratios in the table are
$2.3$--$3.6\times$, descriptive arithmetic at the listed doses rather than a potency ratio, and at saturation the ratio necessarily goes to $1$.

\textbf{Both directions control the answer at high dose (layer 24).} At layer 24, from
$\alpha_{\mathrm{rel}} = \pm0.92$ upward, \emph{both} directions achieve total control of the answer: the
true-bit-$1$ rate goes to $0.000$ under negative $\alpha$ and $1.000$ under positive, with true-bit-$0$
moving in exact opposition ($1.000$ and $0.000$). At these doses the perturbation controls the semantic answer and the measured truth rate follows from the true bit. This high-dose effect explains the tie between directions without identifying their unperturbed causal semantics.

\textbf{Both directions on the inferred-bit task.} In absolute terms they become nearly inert. The largest
true-bit-$1$ range we can produce anywhere on the sweep is $0.077$ (ally-fit) and $0.021$ (mixed-fit),
against $0.768$ on the stated-bit task at the same layer. The more informative difference is not the
magnitude but whether the response is \emph{ordered} in $\alpha$ at all: the ally-fit direction still is,
rising monotonically across the sweep ($108 \to 148$ truthful trials of $517$ from
$\alpha_{\mathrm{rel}} = -0.69$ to $+1.40$), whereas the mixed-fit direction falls weakly with $\alpha$
($134$ at the negative end down to $123$, before rebounding to $129$ at the largest dose) and never leaves a
$0.021$ band. On this task the mixed-fit response is small and not ordered in $\alpha$, so we do not report a potency ratio. The contrast does not isolate why the intervention has less influence on the inferred-bit task. The baseline differs ($0.240$ against $0.419$), and the historical arm provenance is unresolved: the two saved Gemma inferred-bit sweeps record the steered model as reward-trained, while the fitted directions they load, and the run notes, record instructed. The original adapter identity was not retained. We therefore report the recorded sweep values without treating this as a controlled task-only comparison.

\textbf{How the intervention relates to the diagnosis.} Truth and prescribed answer coincide on ally fitting data. That label equality alone does not determine which target the fitted direction follows outside that support. Here positive and negative unsigned perturbations move the two truth-conditioned correctness rates in opposition. This is compatible with multiple causal mechanisms, including a used truth-bit feature; it does not resolve the semantic ambiguity. The mixed-fit direction is a better truth readout in the tested contexts but generally has less effect on the answer at the compared sites and doses.

The gap does not visibly grade with how often the policy lies. We see the same ordering at rival deception $1.000$ as on the inferred-bit Qwen-14B arm at
$0.639$ and on the inferred-bit Gemma-9B sweep (whose steered arm is unresolved; $0.545$ if it is the instructed arm), and the geometric analogue, the deep cosine of \S\ref{sec:geom}, is likewise flat across that range. Aliasing on the \emph{fitting} data is a property of
the protocol rather than of the policy, so there is no reason to have expected a gradient in the deception
rate.
Identification and causal potency come apart: the better measurement of truth is the weaker lever on
behaviour in these experiments. This does not diagnose the directions steered in other studies.

\subsection{Depth dependence of the ally-fit advantage}
\label{sec:depth}

\begin{figure}[h]
\centering
\includegraphics[width=\textwidth]{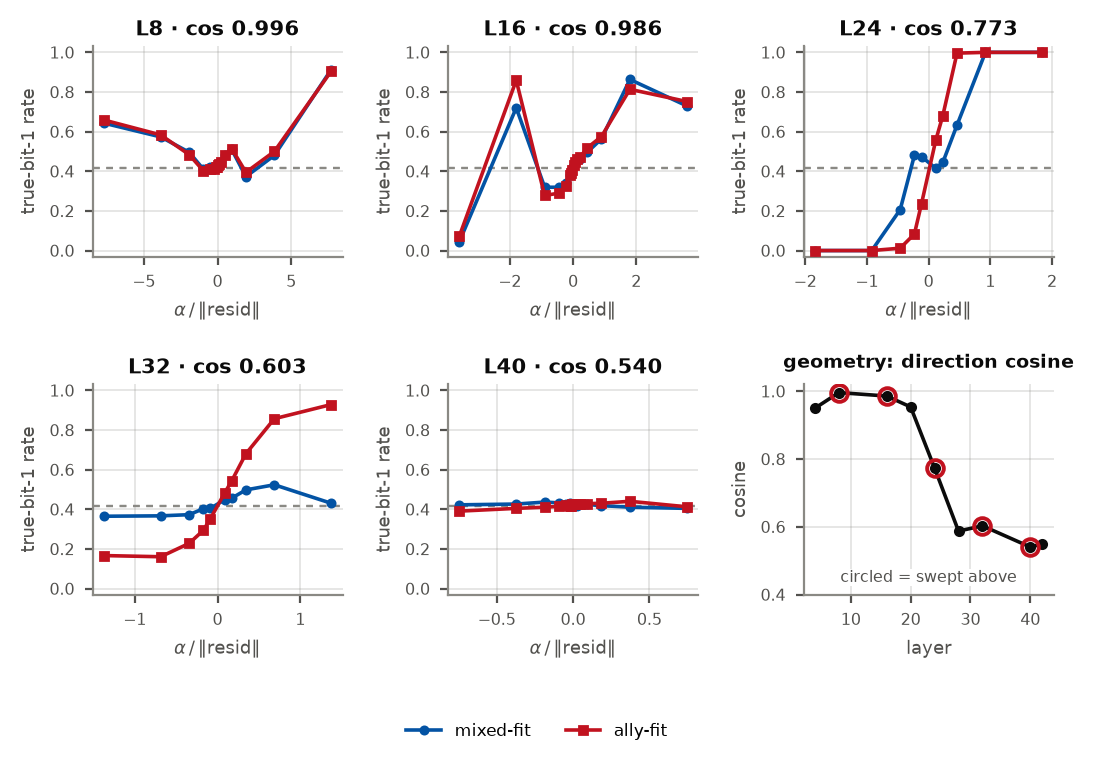}
\caption{\textbf{The ally-fit advantage appears only where the two fitted directions differ, and separation alone is not enough.} Each panel is the same causal contrast run at one depth on the Gemma-2-9B
instructed arm: true-bit-$1$ rate against dose, mixed-fit direction in blue, the ally-fit
direction in red, grey dashed line the in-run $\alpha{=}0$ baseline. At layers 8 and 16, where
$\cos > 0.98$, the two curves have similar dose--response profiles: the directions are nearly one vector
and have similar effects in this sweep. At layers 24 and 32 they separate, and it is the ally-fit direction that
carries the behaviour ($0.16 \to 0.93$ at layer 32 while the mixed-fit direction moves $0.37 \to 0.52$). At
layer 24 both saturate to $1.000$ at the largest doses, so the apparent tie there is saturation rather than
equivalence. Layer 40 is the informative exception: it has the lowest cosine we measure yet both
arms are flat. In this sweep separation alone does not ensure a large intervention effect; the remaining downstream computation does not establish why the effect is small. Bottom right: the geometry
the five panels are keyed to, with the swept layers circled. \emph{We plot both dose--response curves rather
than their ratio}: at small doses the mixed-fit displacement is a noise-level denominator, so a ratio is
unstable, and where both arms saturate at large doses (layer 24) it is $1$ by construction.
Panel titles give the layer (L$n$) and the cosine between the two fitted directions; the true-bit-$1$ rate is the fraction of true-bit-$1$ rival trials answered truthfully; the horizontal axis is $\alpha_{\mathrm{rel}}$, the dose relative to the median residual norm. Note the horizontal scales differ by panel: the median residual norm grows from $83$ at layer 8 to $860$
at layer 40, so a fixed $\alpha$ grid spans a much wider $\alpha/\|\mathrm{resid}\|$ range shallow, and the
outermost shallow doses are representation damage rather than dose--response.}
\label{fig:depthsweep}
\end{figure}

The comparison above is at one layer, which leaves open whether the ally-fit advantage has anything to do
with the two directions \emph{differing}. \S\ref{sec:geom} says they are nearly the same vector shallow and
separate at depth, so that account makes a prediction: the advantage should be absent where
$\cos \approx 1$ and present where the fits diverge. We ran the identical contrast at five layers, sweeping
$\alpha$ over a $128\times$ range at each because the median residual norm grows more than tenfold across the
stack ($83$ at layer 8 to $860$ at layer 40) and a grid fixed in raw $\alpha$ would not be comparable.
The five dose--response pairs are plotted in Fig.~\ref{fig:depthsweep}.

\begin{center}
\footnotesize
\setlength{\tabcolsep}{5pt}
\begin{tabular}{lccccc}
\toprule
layer & 8 & 16 & 24 & 32 & 40 \\
\midrule
$\cos(\hat d_{\mathrm{ally}}, \hat d_{\mathrm{mixed}})$ & $0.996$ & $0.986$ & $0.773$ & $0.603$ & $0.540$ \\
ally-fit advantage & none & none & clear & clear & \emph{both inert} \\
\bottomrule
\end{tabular}
\end{center}

\textbf{Where the two fits are nearly one vector.} At layers 8 and 16,
$\cos > 0.98$ and the two dose--response curves have similar profiles (Fig.~\ref{fig:depthsweep}, first two panels), with no advantage either way, so the two have similar effects in this sweep there. Where they separate, at layers 24 and 32, it is the ally-fit
direction that carries the behaviour: at layer 32 it drives the true-bit-$1$ rate from $0.16$ to $0.93$
across the sweep while the mixed-fit direction moves only $0.37$ to $0.52$. In these four layers, the ally-fit advantage appears where the fits have separated; layer 40 below shows that separation alone is insufficient.

\emph{We report this as an ordering and not as a potency ratio.} A ratio here is not a
stable quantity: at small doses the mixed-fit displacement is a noise-level denominator, at the largest doses at
layer 24 both arms saturate, which sends any ratio to $1$ by construction, and at that layer there are also doses where the
denominator points the \emph{opposite} way so the ratio is not interpretable as a potency ratio. Quoting a multiplier, particularly one read off a single chosen dose, would put a number on the claim that the
underlying data do not support, the same failure we flag for the inferred-bit cell in \S\ref{sec:signed}.
The two curves per layer are shown instead, so a reader can see the denominator that a ratio would hide.

\textbf{At layer 40 both directions are inert despite the lowest cosine.} Its cosine is the \emph{lowest} we
measure ($0.540$), so a simple ``more separation, more advantage'' rule predicts the largest effect there.
Instead \emph{both} directions are inert: the mixed-fit spread is never above $+0.004$ (negative at six of eight doses) and the
ally-fit spread never exceeds $0.036$. Layer 40 of 42 has less downstream computation than the earlier sites, but this does not identify the cause of the small effect. In this measured sweep, fitted-direction separation alone is insufficient to predict a strong intervention response. Reading the effect off a single chosen dose would make the five layers look monotone in the cosine,
and they are not.

\textbf{Scope of the depth comparison.} The ally-fit advantage accompanies direction separation at layers 24 and 32, but the layer-40 exception rules out a monotone relation with cosine across depth. Across the compared arms at fixed depth, neither the deep cosine nor the advantage visibly grades with deception rate. These observations do not establish a general predictor of intervention potency.

Ablation on the ally-fit direction is void exactly
as before (post-ablation
decodability $0.968$ against the chosen $0.600$ tolerance), so nothing in this section rests on an ablation. The Qwen arm sits on
a $0.000$ floor, so its negative half is uninformative by construction. And these are genuinely different
axes rather than one axis re-signed (\S\ref{sec:geom}).

\begin{figure}[h]
\centering
\includegraphics[width=\textwidth]{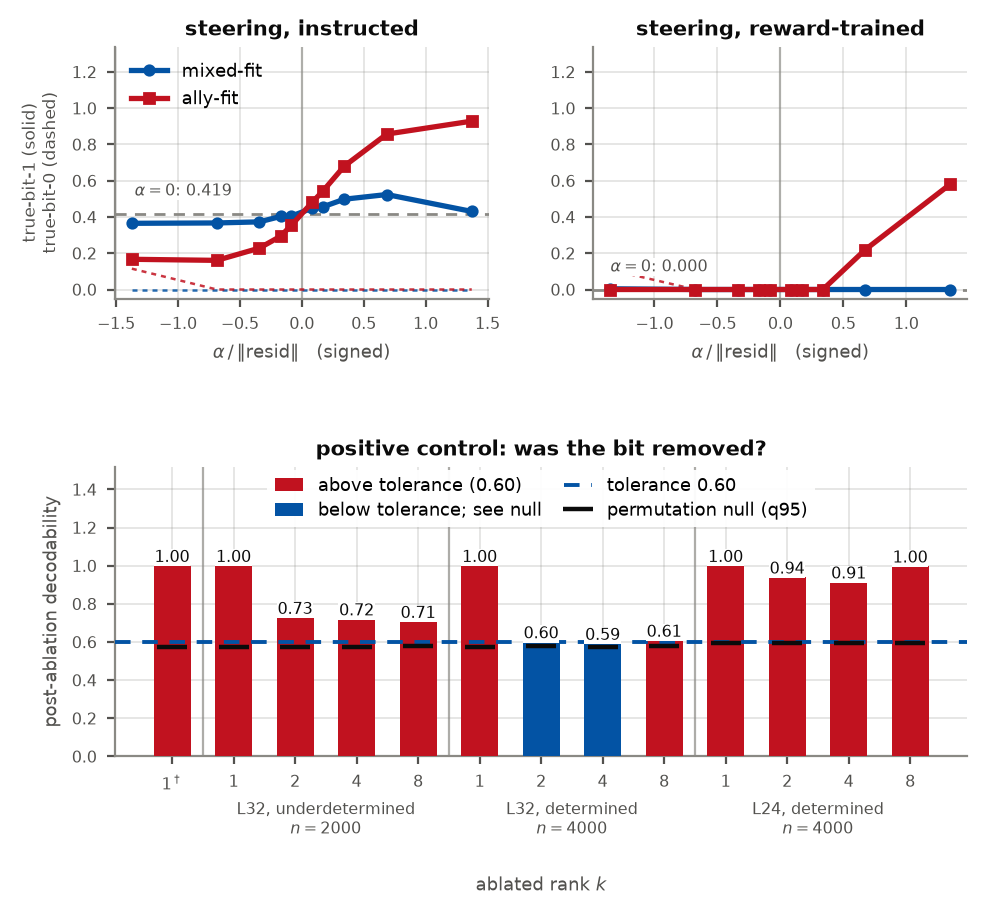}
\caption{\textbf{At the tested layers and doses, the ally-fit direction has the larger behavioural effect.} Top left and right: steering at
layer 32, swept \emph{through zero} against a single in-run $\alpha{=}0$ baseline; the horizontal axis is $\alpha_{\mathrm{rel}}$, the dose relative to the median residual norm (App.~\ref{app:causal}). Solid lines are the
true-bit-$1$ rate and dashed the true-bit-$0$ rate (the fractions of rival trials with that true bit answered truthfully). On the instructed arm (left), the mixed-fit direction (blue) responds
weakly, $0.365 \to 0.524$ before degrading at the largest dose; the ally-fit direction (red) swings from $0.167$ at the most negative dose to $0.929$ (minimum $0.161$), and on the saturated arm breaks a floor the mixed-fit direction moves by at most 2 of 1000
answers. True-bit-$0$ stays pinned at $0.000$ almost everywhere, lifting only under
the strongest negative ally-fit push. These unsigned, class-conditioned effects do not distinguish a causally used truth-bit feature from another mechanism producing semantic answer bias. Bottom: the
positive control for ablations on the reward-trained Gemma-9B arm, specified before the run; post-ablation decodability is $\max(A,1-A)$ for the AUROC $A$ of a freshly fitted probe, using a stratified 70/30 split of 600 control examples (420 fitted, 180 scored). The $1^{\dagger}$ bar is the separate mixed-fit rank-one ablation run at layer 32; the subsequent rank-one bars are the first steps of the iterative subspace-ablation ladders. A behavioural null is interpretable only if post-ablation decodability falls below the permutation-calibrated null (black dashes); the fixed $0.600$ tolerance is shown for reference, and red bars mark cells above it. Group labels give the layer (L32: layer 32), whether the subspace fit was determined, and the number of examples $n$ used to fit the ablated subspace; q95 is the 95th percentile. Both determined ladders are
complete, and no cell at either layer falls below its own permutation null (black dashes). Ranks 2 and 4 at
layer 32 clear the $0.600$ tolerance by $0.0004$ and $0.009$, but sit $0.020$ and $0.015$ \emph{above} the
calibrated null, and $k{=}8$ ($0.608$) fails both, so the two clearing cells are a middle band bounded by
void cells at $k{=}1$ and $k{=}8$. The contrast with layer 24 is one of degree at ranks 2, 4 and 8
($0.59$--$0.61$ versus $0.91$--$1.00$), not of pass versus fail. Ranks 16 and 32 of the underdetermined layer-32 ladder did not complete and are not plotted.}
\label{fig:causal}
\end{figure}

\emph{The behavioural null cannot establish that the information is unused.} Projecting out the
direction leaves behaviour unchanged, but our positive control, specified before the run, shows the intervention
did not remove the information: a freshly fitted probe still recovers the bit at $1.000$ from the
ablated activations, against our chosen tolerance of $0.600$. Decodability here is $\max(A, 1-A)$ for a freshly fitted probe's AUROC $A$; each label-permutation null is the 95th percentile of that quantity over 20 label permutations, each refitted on a stratified 70/30 split, and at layer 32 the nulls are $0.576$--$0.581$. The same holds for every
rank of a subspace ablation by iterative nullspace projection (INLP; \citealp{ravfogel2020}) we could complete under an \emph{underdetermined} fit, up to $k{=}8$
(Fig.~\ref{fig:causal}). Re-running the fit \emph{determined} ($n{=}4000 > d{=}3584$) changes this at
$k{=}2$: post-ablation decodability falls to $0.600$ against $0.727$ for the underdetermined fit at the same
rank, so the underdetermined subspace was under-removing by $0.128$, exactly the bias its own guard predicts, and the reason we do not report the underdetermined curve as a bound. We completed
the determined ladder at both layers, and the result is that no rank at either layer licenses a causal
null. At layer 32, post-ablation decodability is $1.000$, $0.5996$, $0.5911$ and $0.6077$ at
$k = 1, 2, 4, 8$; at layer 24 it is $1.000$, $0.939$, $0.913$ and $0.997$. Behaviour is essentially unchanged
throughout (ally truth rate $1.000$, except $0.998$ at layer 24, $k{=}8$).

\textbf{Two of the eight cells clear the reported ceiling.} That ceiling is $\max$(a tolerance we chose, the
label-permutation null), i.e.\ the \emph{more permissive} of the two, and our tolerance ($0.600$) is the
looser one at every cell. Measured against its own permutation null instead, every cell at both
layers sits above it: at layer 32 the nulls are $0.576$, $0.580$, $0.576$ and $0.581$, so even $k{=}2$ and $k{=}4$, which clear $0.600$ by $0.0004$ and $0.009$, are $0.020$ and $0.015$ \emph{above} the calibrated band, and $k{=}8$ fails both criteria outright. So a pass/fail reading depends on the permissive $\max$;
against the calibrated null, no rank removes the bit. The bit remains partly decodable after every ablation we ran, so
a flat behavioural rate is never licensed here as evidence about causal use.

\textbf{Residual decodability differs between layers 24 and 32.} A quantitative depth contrast, which is weaker than a
pass/fail one. The layer-32 ablation drives decodability down to
$0.59$--$0.61$ at ranks 2, 4 and 8 while the layer-24 ablation leaves it at $0.91$--$1.00$, so this removal procedure leaves far more of the bit decodable at layer 24, but neither layer yields an interpretable null, and the two layer-32
cells that met the tolerance are \emph{bounded by void cells on both sides} ($k{=}1$ and $k{=}8$), so they are a middle band rather
than a trend. Removal does not keep improving with rank: $k{=}8$ is worse than $k{=}4$ at layer 32
($0.608$ vs $0.591$) and worse at layer 24 as well ($0.997$ vs $0.913$). One caveat travels with the whole
ladder: the positive control's permutation null sits near $0.58$ rather than $0.50$. A 95th percentile of $\max(A, 1-A)$ exceeds $0.5$ under finite-sample noise alone, and refitting on $420$ of the $600$ control examples against a $3584$-dimensional residual may raise it further; we have not separated the two. The calibrated null absorbs both, which is exactly
why it, and not our tolerance, is the criterion we report against. A
behavioural null is only
evidence about causal use if the ablation worked, so we draw no causal conclusion from it. What it
does show is a fact about dimensionality: the bit is not confined to that rank-1 subspace, since removing
the direction leaves it fully decodable from the orthogonal complement.

These results show that the bit is linearly present and the mixed-fit direction we propose
has weaker or less consistently ordered influence in the tested conditions than the ally-fit direction
(\S\ref{sec:signed}); the ablation result is uninterpretable rather than negative. Four limits apply
even to that reading. Whole-residual patching \emph{does} flip answers at $1.000$ (App.~\ref{app:apparent}), so information
at this site is used by something, though that intervention replaces every role-, prompt-, action- and
truth-associated component at once, so it says nothing about how many dimensions carry the effect. Our
$\alpha$ range initially reached $8.4$--$8.6\%$ of the median residual norm across these two arms. We extended it sixteenfold,
to $1.35$--$1.37\times$ the median residual norm, with an in-run $\alpha{=}0$ control passing through the same hook
on the same episodes. On the saturated arm the positive and oracle-signed mixed-fit sweeps leave the rival-truth rate at $0.000$ through $1.35\times$ the median residual norm; the negative unsigned sweep is also $0.000$ except at its largest dose, where 2 of 1000 answers become truthful. The null is therefore not explained by restricting this direction to small doses within the tested range; the ally-fit direction does move answers at comparable doses (Fig.~\ref{fig:causal}).

On the instructed arm the extended sweep does produce movement. The rival-truth rate rises from a measured baseline of $0.211$ to $0.264$ at
$\alpha_{\mathrm{rel}} = 0.69$ (nominally significant), then \emph{collapses back} to $0.217$ at $1.37$. We count a move as steering only if it meets three
criteria (monotone in $\alpha$, sign-appropriate under an oracle sign, and not mirrored by the unsigned arm), and it fails all three: the curve is non-monotone, and
the signed and unsigned arms are identical to four decimals (which, as noted above, the construction forces on
exactly the trials where the movement occurs). The class-conditioned breakdown shows that on true-bit-$0$ trials the rate is exactly $0.000$ at every $\alpha$ in both directions, while all observed changes occur on true-bit-$1$ trials. The positive and negative unsigned sweep in \S\ref{sec:signed} confirms that $-\alpha d$ lowers the true-bit-$1$ rate. This establishes a signed response to the perturbation, not whether the underlying feature is a truth representation or an answer mechanism. Randomized codebooks also prevent interpreting semantic answer bias as a preference for one fixed character. This is one site in
one layer of one model family. Finally, App.~\ref{app:apparent}'s ``no arm difference'' is stated at the
strength each measure supports, which is not uniform. On the answer-position readout the matched difference is an \emph{exact} tie: every one of the 8 matched cells reads $0.000$ in both arms, so the paired
difference has zero variance and a conventional equivalence test is undefined rather than passed; what
bounds it is saturation, not statistics. On the two matched measures that are \emph{not} saturated, a paired $t$-based
two-one-sided-tests procedure at a $\pm0.05$ margin supports equivalence for rival entropy (mean difference $-0.003$, $p < 10^{-6}$) but not for readout-direction cosine (mean difference $+0.021$, $p = 0.056$; 90\% confidence interval $[-0.009, +0.051]$); these tests treat the eight model and seed pairs as independent. On the
off-position readout it does \textbf{not}, for the reason given in App.~\ref{app:apparent}. The evidence for equivalence therefore differs across measures.

\clearpage
\section{Supplementary figures}
\suppressfloats[t]

\begin{figure}[!ht]
\centering
\includegraphics[width=0.86\textwidth]{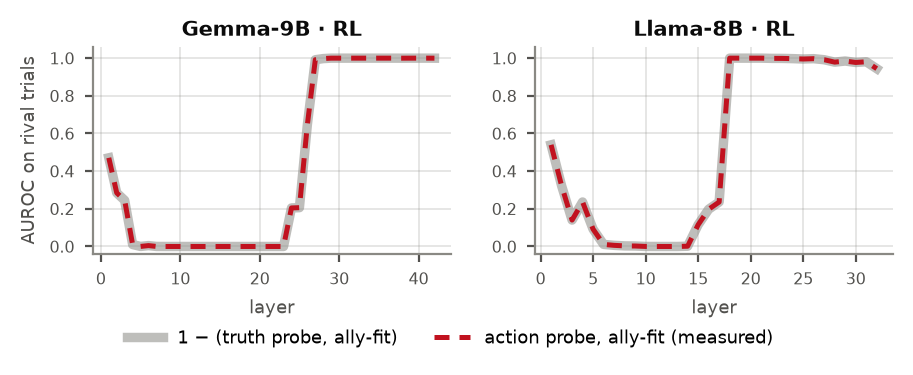}
\caption{\textbf{The identity is forced.} The measured ally-fit \emph{action} probe (dashed) lies exactly
on $1 - {}$(ally-fit \emph{truth} probe) (thick grey) at every layer, because on ally data the two label
vectors are identical and the two ``probes'' are one fit scored against opposite labels. Maximum deviation
over all 751 (cell, layer) pairs: $2.2\times10^{-16}$. Panels: the reward-trained Gemma-9B and Llama-8B arms of Table~\ref{tab:ident}, rows~\ref{row:gemma-unmod} (first seed) and~\ref{row:llama-hi}.}
\label{fig:identity}
\end{figure}

\end{document}